\documentclass[]{ustc_conference}
\usepackage{amsmath}
\usepackage{amsthm}
\usepackage{amssymb}
\usepackage[utf8]{inputenc}
\usepackage{booktabs}
\usepackage{multirow}
\usepackage[table]{xcolor}
\usepackage{adjustbox}
\usepackage{wrapfig}
\usepackage[edges]{forest}
\usepackage{tikz}
\usepackage{amsfonts}
\usepackage[toc,page,header]{appendix}
\usepackage{bbm}
\usepackage{enumitem}
\usepackage{mathtools}
\renewcommand{\paragraph}[1]{\vspace{0.1em}\noindent\textbf{#1}}

\definecolor{frameorange}{RGB}{218,114,27}
\definecolor{bgorange}{RGB}{253,243,235}
\definecolor{frameblue}{RGB}{0,85,150}
\definecolor{bgblue}{RGB}{235,245,255}
\definecolor{framegreen}{RGB}{34,139,34}
\definecolor{bggreen}{RGB}{240,253,240}
\newtcolorbox{StrategyBox}[3][frameorange]{
  enhanced, breakable, width=\textwidth,
  title={#3}, colframe=#1, colback=#2, colbacktitle=#1, coltitle=white,
  fonttitle=\bfseries\large, fontupper=\rmfamily, arc=1.5mm, boxrule=1.2pt,
  top=3mm, bottom=3mm, left=3mm, right=3mm,
  toptitle=0.5mm, bottomtitle=0.5mm
}
\newcounter{promptidx}
\newcounter{caseidx}

\newcommand{\Method}{Mind2Report}
\newcommand{\Evaluation}{QRC-Eval}

\title{\Method: Expert-Level Commercial Report Synthesis via Cognitive Deep Research Agent}

\author{%
\parbox{\textwidth}{\centering
Mingyue Cheng\frontsup{1}, Daoyu Wang\frontsup{1}, Qi Liu\frontsup{1\corrauthor}, Shuo Yu\frontsup{1}, Xiaoyu Tao\frontsup{1},\\
Yuqian Wang\frontsup{1}, Chengzhong Chu\frontsup{2}, Yu Duan\frontsup{2}, Mingkang Long\frontsup{2}, Enhong Chen\frontsup{1}
}}

\affiliation{%
\parbox{\textwidth}{\centering
\affilsup{1}State Key Laboratory of Cognitive Intelligence, University of Science and Technology of China \\
\affilsup{2}Artificial Intelligence Engineering Institute, iFLYTEK Co., Ltd
}}

\correspondence{\email{daoyu.wang@mail.ustc.edu.cn}, \email{mycheng@ustc.edu.cn}, \email{qiliuql@ustc.edu.cn}}

\abstract{
Synthesizing informative commercial reports from massive and noisy web sources is critical for high-stakes business decisions. Although recent deep research agents (DRAs) achieve notable progress, their reports remain limited in quality, reliability, and coverage. These mainly stem from ambiguous intents that cause search drift, retrieved web content that rapidly saturates the context window, and single-pass synthesis that limits report comprehensiveness. In this work, we propose \Method, a cognitive deep research agent that emulates commercial analysts to synthesize expert-level reports. \Method\ first probes fine-grained commercial intent to establish a structured outline, then recursively explores web sources and distills validated evidence into research memory to preserve context efficiency. Meanwhile, the research memory and outline continuously co-evolve, refining the report structure to avoid rigid initial planning. Finally, \Method\ iteratively synthesizes the report based on the evolving outline and accumulated evidence. Together, these designs enable reliable and context-efficient long-horizon commercial deep research. To rigorously evaluate commercial DRAs, we further construct \Evaluation, comprising 200 real-world commercial tasks and a holistic evaluation framework covering report quality, reliability, and coverage. Extensive experiments demonstrate that \Method\ consistently outperforms leading proprietary and open-source DRAs, while ablations verify the effectiveness of each module and further analyze the challenges they address. We expect this work to advance the development of commercial deep research agents.

}

\checkdata[Publication]{Accepted by CIKM '26, Rome, Italy}
\checkdata[Code and Data]{\url{https://github.com/ustc-ai4science/Mind2Report}}

\begin{document}

\maketitle

\section{Introduction}


Synthesizing informative commercial reports, such as competitor analysis and market assessments, from massive and noisy web sources underpins high-stakes business decisions \cite{shiller2003irrationalexuberance,zhang2025xfinbench}. In practice, a professional commercial analyst typically needs to clarify imprecise requirements, record key evidence and draft structured reports, which is a labor-intensive process \cite{dong2025large,liu2025real}. Consequently, automated commercial report synthesis emerges as a critical task, garnering extensive research attention \cite{le2025rag,xu2025comprehensive}.

Researchers initially approached commercial report synthesis through statistical text extraction methods, restricting the task to basic short-form summarization \cite{dagdelen2024structured}. Fortunately, the rise of large language models (LLMs) unlocks the potential for long-form commercial report synthesis, which requires strategic analysis over massive, time-sensitive, and noisy web information collected from diverse sources \cite{wang2025paperarena,yao2025rigorous}. While retrieval-augmented generation (RAG) facilitates single-pass synthesis, the static retrieval-generation pipeline often limits information coverage and suffers from retrieval noise \cite{sun2025redeep,yu2025multi}. More recently, deep research agents (DRAs) significantly advance commercial report synthesis by coupling autonomous planning with multi-step tool invocation \cite{openai2025deepresearch,team2025tongyi}. These agents iteratively refine search objectives, verify evidence across heterogeneous sources, and organize long-horizon information, thereby emulating the rigorous workflow of professional commercial analysts \cite{team2025mirothinker}.

\begin{wrapfigure}{r}{0.52\linewidth}
    \centering
    \includegraphics[width=\linewidth]{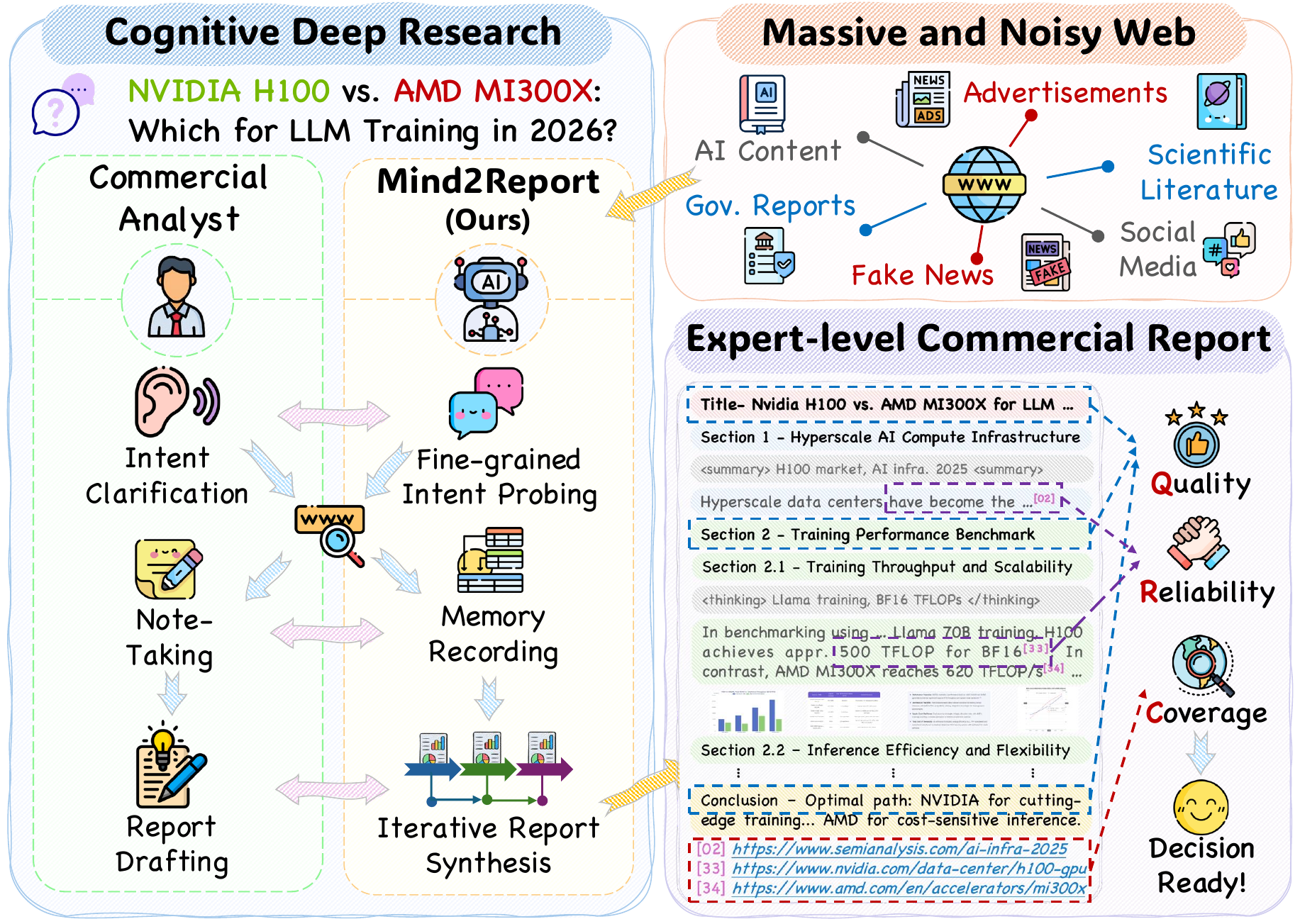}
    \captionof{figure}{\Method~emulates a professional commercial analyst to synthesize expert-level commercial reports from massive and noisy web sources.}
    \label{fig:1}
\end{wrapfigure}

Despite their strong progress, current DRAs still exhibit unresolved limitations in commercial report synthesis, particularly in quality, reliability, and coverage \cite{openai2025deepresearch,team2025tongyi,gu-etal-2025-rapid}. In practice, these limitations mainly stem from three challenges. First, commercial analysis queries are often broad and underspecified, which easily causes search drift during long-horizon research \cite{gu-etal-2025-rapid,team2025mirothinker}. Second, retrieved web content rapidly saturates the context window with noisy and redundant information, which weakens reasoning reliability and can eventually lead to hallucinated reports \cite{sun2025redeep,yu2025multi}. Third, single-pass synthesis often underuses the collected evidence, resulting in reports with limited information richness and insufficient analytical depth \cite{yao2025rigorous,sun2025redeep}. As illustrated in Figure \ref{fig:1}, these motivate us to develop a commercial deep research agent that follows the rigorous workflow of professional analysts, from requirement clarification and evidence management to comprehensive report synthesis.

To address these challenges, we propose \Method, a cognitive deep research agent that emulates professional commercial analysts to synthesize expert-level commercial reports. To mitigate search drift caused by ambiguous commercial queries, \Method\ first probes fine-grained commercial intent to establish a structured report outline. It then recursively explores web sources and distills validated evidence with source citations into research memory, which preserves context efficiency during long-horizon research under massive and noisy web environments. Meanwhile, the research memory and outline continuously co-evolve, enabling adaptive refinement of the report structure to avoid rigid initial planning. Finally, \Method\ iteratively synthesizes the report based on the evolving outline and accumulated evidence in the research memory, while maintaining grounded citation sources for evidence tracing.

Furthermore, we propose \Evaluation\ to systematically evaluate long-form commercial reports in a method-independent manner. It comprises 200 time-sensitive commercial queries manually crafted by commercial experts to ensure high quality and practical relevance. To reduce temporal variation under dynamic web environments, we cache retrieved web content during evaluation to improve reproducibility across different methods. We further establish a holistic evaluation framework covering report quality, reliability, and coverage, where each dimension contains specialized metrics designed to capture critical aspects of commercial reports. Extensive experiments demonstrate that \Method\ consistently outperforms leading proprietary and open-source DRAs, including OpenAI and Gemini DRAs \cite{openai2025deepresearch,google2024deepresearch}. Detailed ablation studies further verify the effectiveness of each module and analyze the challenges they address. Moreover, we verify the alignment between our proposed metrics and human expert judgment, demonstrating the rationality of \Evaluation\ metrics for commercial deep research assessment. Overall, we hope \Method\ and \Evaluation\ can contribute to the future development of commercial deep research agents and evaluation strategies.

\newpage

Our contributions can be summarized as follows:
\begin{itemize}
\item We propose \Method, a cognitive deep research agent for expert-level commercial report synthesis, which mitigates search drift, noisy retrieval, and limited evidence utilization under massive web environments through intent-aware planning, validated evidence memory, and iterative outline refinement.
\item We construct \Evaluation, comprising 200 real-world commercial queries and a holistic evaluation framework for systematically assessing quality, reliability, and coverage across synthesized commercial reports with cached web evidence, expert-crafted queries, and dimension-specific automatic metrics.
\item We conduct extensive experiments and detailed analyses, demonstrating the effectiveness of \Method\ against leading proprietary and open-source deep research agents in long-form commercial report synthesis, including module ablations, human-alignment studies, and fine-grained domain comparisons.
\end{itemize}

\section{Related Work}

This section reviews progress in automated report synthesis, summarizes the main directions of deep research agents, and finally highlights the distinctive features of \Method.

\subsection{Automated Report Synthesis}

Early research frames automated report synthesis as a text summarization task, using statistical or extractive methods to identify salient sentences from source documents \cite{sundaram2023automating,liu-etal-2023-evaluating}. The emergence of LLMs shifts this task from short-form extraction \cite{achiam2023gpt} to long-form generative synthesis \cite{bai2025longwriter}, enabling models to produce more structured and coherent reports \cite{lee-etal-2025-navigating}. To improve factual grounding and information access, researchers further introduce retrieval-augmented generation (RAG), which allows LLMs to incorporate external knowledge during generation \cite{cheng2025survey,gao2023retrieval}. Recent works also explore graph-based retrieval \cite{edge2024local}, self-reflective retrieval \cite{asai2024self}, and evidence-grounded generation \cite{gao2023enabling,sorodoc2025garage} to improve reasoning quality and citation traceability \cite{sun2025enhancing,ouyang2025hoh}. Based on these advances, two representative directions have received increasing attention. One focuses on academic survey generation \cite{wang2024autosurvey}, where systems organize evidence from scientific literature or web sources into structured long-form articles \cite{shao2024assisting,susnjak2025automating}. The other focuses on financial and commercial report synthesis \cite{le2025rag,hu2026finsearchcomp}, where LLMs analyze domain-specific information to support complex business analysis and investment decisions \cite{zhang2025xfinbench,xu2025comprehensive}. Unlike prior systems that primarily retrieve and summarize information in a fixed pipeline, \Method treats commercial report synthesis as a long-horizon process that jointly evolves user intent, evidence memory, and report structure throughout the research lifecycle.

\subsection{Deep Research Agents}

Deep research agents (DRAs) extend LLM-based systems from single-pass generation to autonomous long-horizon research \cite{xu2025comprehensive,openai2025deepresearch,wu2026webdancer}. Modern DRAs combine planning \cite{zhang2025far}, query refinement \cite{team2025mirothinker}, web exploration \cite{team2025tongyi}, tool invocation \cite{li2025webthinker}, evidence verification \cite{zheng2025deepresearcher}, and report generation to complete open-ended research tasks \cite{openmanus2025}. Existing approaches follow two directions. Training-based methods typically use reinforcement learning to improve multi-step reasoning and tool use \cite{cheng2025agent}, but often require complex reward design and substantial training cost \cite{jiang2026tablemind,zheng2025deepresearcher}. Workflow-based methods instead orchestrate strong base LLMs with tools, memory, and context management, offering more flexibility \cite{lu2024ai,miroflow2025framework}. Meanwhile, recent evaluation frameworks have begun to move beyond lexical metrics \cite{papineni2002bleu} and assess deep research agents from multiple perspectives \cite{du2026deepresearch,samarinas-etal-2025-beyond,java2026characterizing,sun2025finresearchbench}. 
Despite this progress, existing DRAs and evaluations mainly target general research scenarios, leaving high-stakes commercial report synthesis less explored. Commercial deep research faces distinct challenges, including search drift from broad intents, context saturation from noisy web information, and evidence underuse in single-pass synthesis. Different from generic DRA workflows, \Method\ combines commercial intent probing, research memory, and memory-outline co-evolution for context-efficient long-horizon synthesis. \Evaluation\ further assesses commercial DRAs across quality, reliability, and coverage.

\begin{figure*}[t]
    \centering
    \includegraphics[width=1\textwidth]{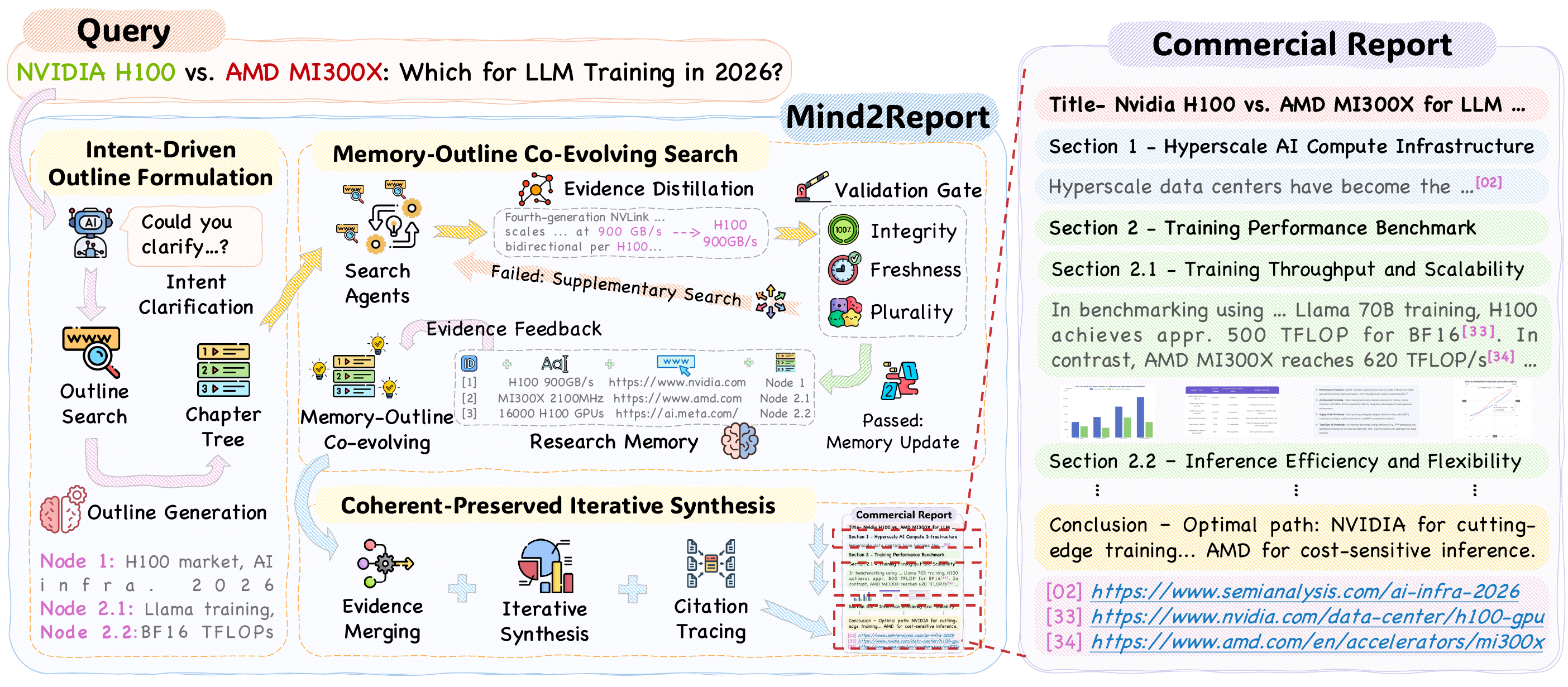}
    \caption{The framework of \Method. Given a commercial query, \Method\ includes three key components: intent-driven outline formulation, memory-outline co-evolving search and coherent-preserved iterative synthesis.}
    \label{fig:2}
\end{figure*}

\section{\Method}

In this section, we first formalize the commercial deep research problem and then present the overall design of \Method\ for expert-level report synthesis, covering its core components and their roles in report synthesis.

\subsection{Problem Definition}
Commercial deep research requires an autonomous agent to synthesize a long-form report from massive and noisy web sources according to an open-ended business query.
Formally, given an initial query $Q$, the agent interacts with a web environment over a finite horizon $T$.
At step $t$, the agent maintains an internal research state $s_t=(q_t,O_t,M_t,C_t)$, where $q_t$ denotes the refined task intent, $O_t$ denotes the evolving report outline, $M_t$ denotes the research memory containing validated evidence, and $C_t$ denotes the citation source registry associated with the evidence.
Based on $s_t$, the agent performs an action $a_t$, such as search, evidence extraction, or synthesis, and receives an observation from external sources.
The objective is to update the research state through these interactions and finally produce a commercial report $R$ whose factual claims are supported by traceable citation sources.

\subsection{Overview of \Method}
Figure~\ref{fig:2} illustrates the framework of \Method\ for synthesizing commercial reports from an initial query $Q$.
The key design is to maintain an evidence-centered research state throughout the deep research process.
First, \Method\ probes fine-grained commercial intent and constructs an initial outline that converts an underspecified request into a structured analytical scaffold.
Second, guided by the outline, \Method\ invokes specialized search agents to recursively explore web sources and distill observations into validated evidence stored in research memory.
Each memory entry preserves compact evidence content, its citation source, and its association with the corresponding outline node.
Meanwhile, the research memory and outline continuously co-evolve: the outline constrains evidence acquisition, while accumulated memory evidence updates the evidence state of each outline node and exposes section-level gaps for subsequent search.
Finally, \Method\ synthesizes the report section by section from the evolving outline and its associated memory evidence, thereby improving evidence utilization while preserving citation tracing.

\subsection{Intent-Driven Outline Formulation}
Commercial queries are often broad and underspecified, which can cause long-horizon research to drift toward loosely related topics.
To mitigate this issue, \Method\ begins with intent-driven outline formulation.
Given the initial query $Q$, the agent first clarifies the commercial intent behind the request, including the target domain, analysis objective, expected granularity, temporal scope, and key dimensions of interest.
The clarified intent forms a refined task specification $q_0$ that constrains the subsequent research process.
Guided by $q_0$, \Method\ conducts a preliminary search to collect essential background information and then organizes the retrieved information into an initial outline $O_0$.
This outline represents the query as a structured chapter tree, where each section corresponds to a specific analytical focus rather than a generic writing slot.
By converting an ambiguous request into an explicit analytical scaffold, \Method\ reduces search drift and provides a coherent roadmap for evidence acquisition and report synthesis.
Importantly, $O_0$ is treated as an initial scaffold instead of a fixed plan, since commercial evidence may reveal missing dimensions during later research; each section in the outline can be viewed as an outline node $o_j \in O_t$.

\subsection{Memory-Outline Co-Evolving Search}
After obtaining the initial outline, \Method\ conducts iterative exploration through specialized search agents. For each outline node $o_j \in O_t$, a search agent queries heterogeneous web sources using the node-level requirement and current memory slice $M_t(o_j)$, and distills retrieved pages into candidate evidence. \Method\ evaluates this evidence through three binary gates: integrity, freshness, and plurality, which respectively assess factual support, topic-dependent temporal validity, and sufficient coverage of required examples or perspectives. If any activated gate fails, the agent generates targeted supplementary queries and continues exploration until all gates pass or the search horizon is reached. Only validated evidence is stored in research memory as $m_i=(e_i,c_i,o_j)$, where $e_i$ is a compact evidence statement, $c_i$ the citation source, and $o_j$ the associated outline node. This interface lets the main agent interact with search primarily through $M_t$ and $O_t$, reducing context saturation while preserving citation tracing.

Research memory further enables the outline to evolve with accumulated evidence, distinguishing \Method\ from query-level search refinement. After validated entries ${m_i}$ are returned for $o_j$, \Method\ updates $M_{t+1}=M_t \cup {m_i}$ and $O_{t+1}=\mathrm{Update}(O_t,M_{t+1})$. Memory entries sharing the same outline node and citation sources are also merged before synthesis to reduce fragmentation. Consequently, each outline node is progressively enriched with supporting evidence and remaining gaps rather than remaining a static plan. Subsequent search is conditioned on both $O_{t+1}$ and $M_{t+1}(o_j)$: supported aspects can be reused for synthesis, while uncovered aspects trigger targeted exploration. In this way, the outline guides evidence acquisition, while accumulated memory refines subsequent search and writing, mitigating rigid planning and transforming fragmented web observations into reusable section-level support and improving overall reasoning coherence across iterative cycles.

\subsection{Coherent-Preserved Iterative Synthesis}
\Method\ produces the final commercial report through coherent-preserved iterative synthesis with citation tracing.
Given the final outline $O_T$, research memory $M_T$, and citation source registry $C_T$, the agent generates the report section by section following the outline hierarchy.
For each section, \Method\ retrieves the associated memory evidence and consolidates evidence supported by the same citation sources into coherent analytical support.
The preceding report context is used to preserve discourse continuity, while factual statements remain grounded in section-level memory rather than the entire search trace.
This allows the agent to synthesize each segment under a compact local evidence context and improves the reuse of collected evidence.
Citation placeholders are then matched back to the source registry $C_T$ maintained in the research state, producing a final report $R$ with traceable citation sources.
This process mitigates the limitations of single-pass generation by preserving cross-section coherence, reducing evidence underuse, and supporting final report citation tracing.

\section{\Evaluation}

In this section, we detail the construction of \Evaluation, its key features, and the multi-dimensional automatic evaluation strategy for assessing agent capabilities.

\subsection{Dataset Construction}
\begin{wrapfigure}{r}{0.52\linewidth}
    \centering
    \includegraphics[width=\linewidth]{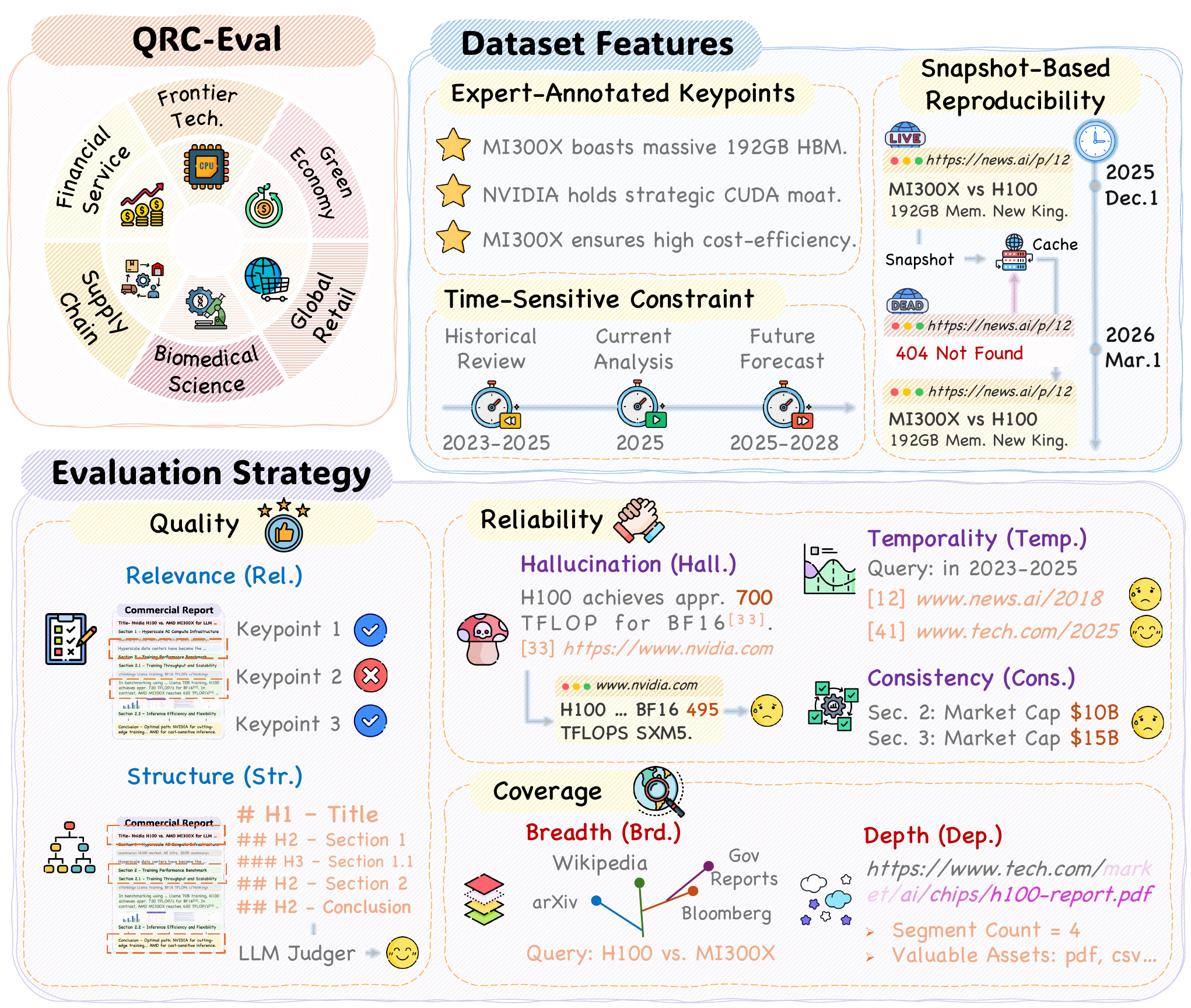}
    \captionof{figure}{Overview of the \Evaluation, a commercial query suite and a holistic evaluation strategy designed for assessing commercial reports via quality, reliability, and coverage.}
    \label{fig:3}
\end{wrapfigure}

As shown in Figure~\ref{fig:3}, we construct a dataset comprising 200 time-sensitive commercial queries manually crafted by business experts to ensure professional quality.
The design process incorporates complex analytic intents to simulate real-world business scenarios.
To evaluate generalization capabilities across diverse commercial scenarios, we distribute these queries roughly evenly among six distinct commercial domains.
Furthermore, this manual creation process ensures an unbiased assessment for all methods.
Detailed construction and data distribution appear in Appendix~\ref{app:a}.

\subsection{Dataset Key Features}
The dataset exhibits three distinctive features designed to address the unique challenges of commercial research.
First, we utilize keypoints annotated by experts to serve as a reference.
Experts identify critical information dimensions such as technical specifications and strategic market positions for each query.
Second, the dataset enforces strict temporal constraints across the queries.
We categorize tasks into historical reviews, current analyses, and future forecasts to assess how agents handle temporal information dynamics.
This design challenges models to distinguish between outdated context and recent developments efficiently.
Third, we adopt a reproducibility-first design based on fixed queries, expert keypoints, source timestamps, and web snapshots to address the volatility of online information.
Since web content frequently changes or becomes inaccessible over time, we cache the exact state of citation sources at the time of our experiments.
This frozen retrieval corpus guarantees that all methods interact with consistent environments, enabling later re-evaluation without being confounded by source updates or link decay.
We also release the complete retrieval corpus together to support reproducible comparison.

\subsection{Multi-Dimensional Evaluation Strategy}

We formalize the final report as an ordered sequence of claim-source pairs, so \Evaluation\ evaluates final reports and citation sources, avoiding assumptions about specific agent workflows.
The quality dimension evaluates content relevance by measuring the alignment between claims and keypoints.
We also assess the structure via hierarchical headers to ensure the logical coherence.
Reliability ensures trustworthiness through the hallucination rate which penalizes claims that lack support from citation sources.
We further measure temporality by verifying that source timestamps satisfy the temporal constraints and evaluate consistency by detecting numerical or logical contradictions across the context.
Coverage includes source breadth, which quantifies information diversity, and search depth, which evaluates path segments of retrieved sources.
Given the potential biases of LLM-as-a-judge evaluation~\cite{zheng2023judging}, we prioritize standard or deterministic calculations whenever objective signals are available, such as keypoint matching, timestamp verification, URL accessibility, source-domain statistics, and path-based depth.
We only employ LLM-as-a-judge for semantic judgments that are difficult to calculate with rule-based metrics, such as structure (Str.), hallucination (Hall.), and consistency (Cons.).
Additionally, we track profile metrics including report length (Len.) and processing time (Time). These serve as references and do not influence the final ranking.
Detailed metrics formulas appear in Appendix~\ref{app:b}.

\begin{table*}[t]
    \centering
    \caption{Performance of \Method\ against leading baselines through \Evaluation. Key metrics include relevance (Rel.), structure (Str.), hallucination (Hall.), temporality (Temp.), consistency (Cons.), normalized breadth (Brd.), normalized depth (Dep.), report length (Len.), and time (Time). \textbf{Bold} means the best and \underline{underline} is the second best.}
    \label{tab:1}
    \resizebox{0.99\textwidth}{!}{%
    \begin{tabular}{p{5.2cm}ccccccccccc}
        \toprule
        \multirow{2}{*}{\textbf{Methods}} & 
        \multicolumn{2}{c}{\textbf{Quality}} & 
        \multicolumn{3}{c}{\textbf{Reliability}} & 
        \multicolumn{2}{c}{\textbf{Coverage}} & 
        {\textbf{Avg.}} & 
        \multicolumn{2}{c}{\textbf{Profile}} \\
        \cmidrule(lr){2-3} \cmidrule(lr){4-6} \cmidrule(lr){7-8} \cmidrule(lr){10-11}
         & Rel.~$\uparrow$ & Str.~$\uparrow$ & Hall.~$\downarrow$ & Temp.~$\uparrow$ & Cons.~$\uparrow$ & Brd.~$\uparrow$ & Dep.~$\uparrow$ & \textbf{Rank} & Len. & Time \\
        \midrule
        \multicolumn{11}{l}{\textit{Proprietary DRAs}} \\
        \hspace{1.5em}o3 Deep Research & 63.52 & \underline{79.18} & \underline{10.48} & 79.85 & \underline{64.13} & \underline{70.80} & 81.75 & \underline{2.43} & 38.34k & 516s \\
        \hspace{1.5em}o4-mini Deep Research & 54.23 & 72.09 & 16.54 & 70.47 & 48.21 & 40.35 & 68.25 & 6.43 & 12.62k & 364s \\
        \hspace{1.5em}Gemini Deep Research & \underline{64.87} & 78.54 & 11.25 & \underline{81.23} & 63.58 & 66.35 & \underline{83.75} & 2.71 & 46.91k & 498s \\
        \hspace{1.5em}Grok Deep Search & 59.54 & 75.39 & 13.76 & 76.52 & 55.45 & 66.85 & 57.50 & 4.86 & 13.15k & 127s \\
        \hspace{1.5em}Perplexity Deep Research & 58.17 & 71.53 & 15.22 & 78.41 & 52.86 & 32.85 & 52.75 & 7.00 & 18.43k & 229s \\
        \arrayrulecolor{gray!70}\cmidrule(lr){1-11}\arrayrulecolor{black}
        \multicolumn{11}{l}{\textit{Open-source Training-based DRAs}} \\
        \hspace{1.5em}WebThinker & 49.53 & 66.18 & 19.47 & 66.85 & 42.54 & 52.95 & 61.00 & 8.43 & 5.34k & 263s \\
        \hspace{1.5em}MiroThinker & 52.84 & 68.52 & 18.23 & 69.48 & 45.19 & 39.45 & 53.50 & 8.57 & 6.58k & 315s \\
        \hspace{1.5em}Tongyi-DeepResearch & 55.46 & 70.25 & 17.58 & 72.43 & 49.87 & 44.50 & 79.50 & 6.14 & 9.84k & 624s \\
        \arrayrulecolor{gray!70}\cmidrule(lr){1-11}\arrayrulecolor{black}
        \multicolumn{11}{l}{\textit{Open-source Workflow-based DRAs}} \\
        \hspace{1.5em}MiroFlow & 46.52 & 62.84 & 23.47 & 63.45 & 36.53 & 38.50 & 62.75 & 10.29 & 3.58k & 262s \\
        \hspace{1.5em}OpenManus & 48.25 & 65.14 & 21.82 & 65.23 & 39.38 & 48.95 & 65.00 & 8.86 & 5.86k & 146s \\
        \hspace{1.5em}OWL & 44.56 & 60.52 & 24.58 & 61.54 & 33.25 & 34.30 & 63.50 & 11.14 & 9.58k & 287s \\
        \midrule
        \textbf{\Method} (Ours) & \textbf{75.42} & \textbf{85.24} & \textbf{6.12} & \textbf{90.53} & \textbf{75.82} & \textbf{80.85} & \textbf{84.25} & \textbf{1.00} & 21.93k & 385s \\
        \bottomrule
    \end{tabular}%
    }
\end{table*}

\section{Experiments}
In this section, we first introduce the experimental setup, then report the main results and ablation studies of \Method.
We further present detailed case studies and failure analysis, and finally validate the alignment between \Evaluation\ and human judgment.

\subsection{Experimental Setup}

\paragraph{Baselines.}
We evaluate \Method\ against a comprehensive set of baselines categorized into three distinct groups. The first group encompasses proprietary deep research agents, including o3 Deep Research \cite{openai2025deepresearch}, o4-mini Deep Research \cite{openai2025deepresearch}, Gemini Deep Research \cite{google2024deepresearch}, Grok Deep Search, and Perplexity Deep Research. The second group comprises open-source training-based DRAs, specifically WebThinker \cite{li2025webthinker}, MiroThinker \cite{team2025mirothinker}, and Tongyi-DeepResearch \cite{team2025tongyi}. Finally, we compare against open-source workflow-based DRAs that orchestrate LLMs and external tools for deep research tasks, including MiroFlow \cite{miroflow2025framework}, OpenManus \cite{openmanus2025}, and OWL \cite{hu2025owl}.
For each baseline, we use its official API, released checkpoint, or official repository, and follow the recommended settings whenever available.
For methods without explicit query rewriting or expansion, we apply the same intent probing algorithm as \Method\ to reformulate the query for a fairer comparison.

\paragraph{Implementation Details.}
We employ DeepSeek-V3.1~\cite{liu2024deepseek} as the backbone LLM and DeepSeek-R1~\cite{guo2025deepseek} specifically for outline generation. To ensure fairness, we standardize inference parameters across all LLMs. We equip all methods with identical Google Search tools, except proprietary DRAs whose internal tools are inaccessible. For reproducibility, cached webpage snapshots are used preferentially when available. To ensure stability, we execute three independent runs for each method and report the average metrics. For all LLM-as-a-Judge scenarios, we adopt Gemini 2.5 Pro~\cite{google2024deepresearch} as the unified evaluator. Detailed parameters are provided in Appendix~\ref{app:d}.

\newpage

\subsection{Main Results}
\begin{wrapfigure}{r}{0.52\linewidth}
    \centering
    \includegraphics[width=\linewidth]{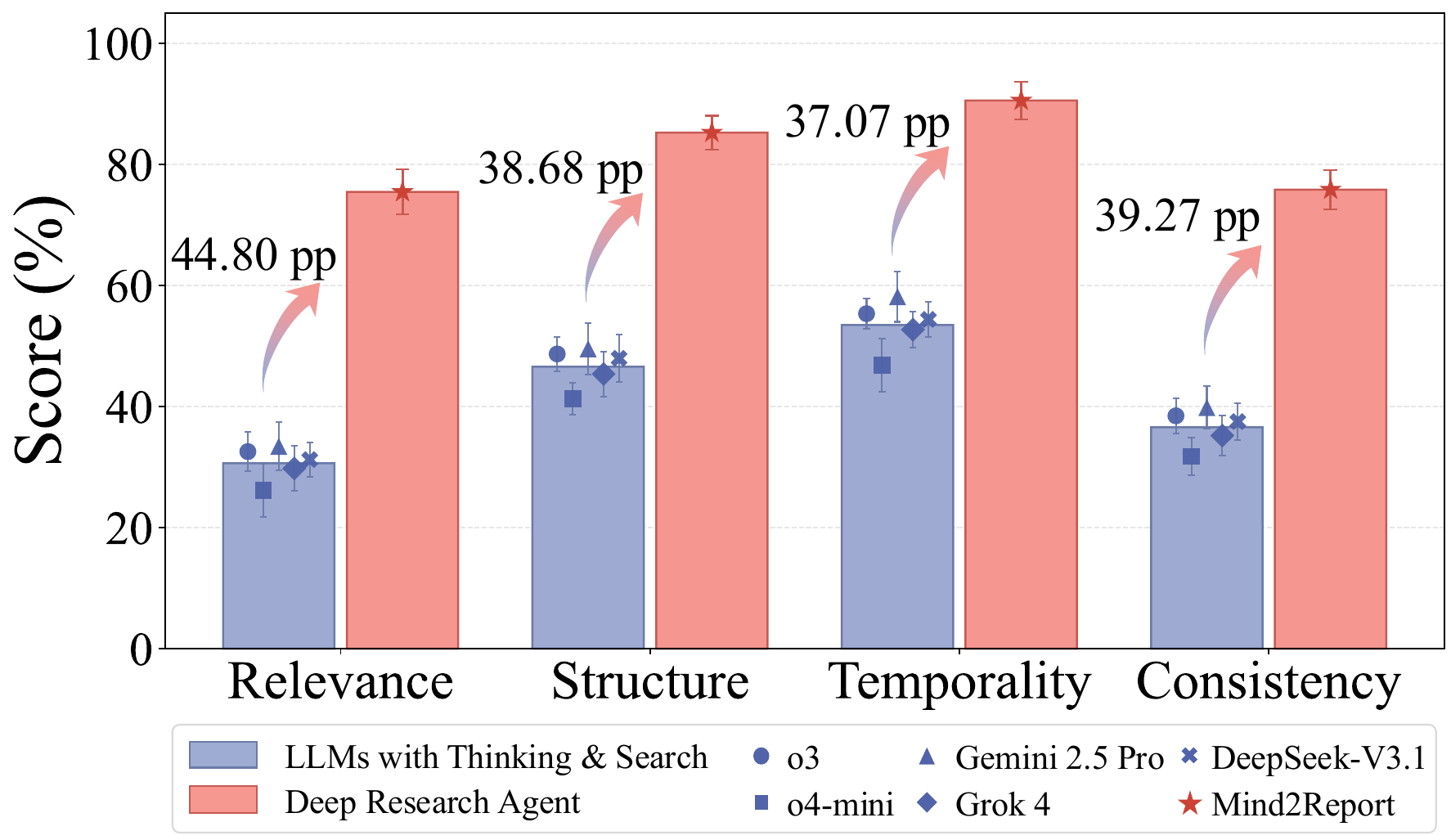}
    \captionof{figure}{\Method~significantly outperforms leading LLMs  with thinking and search, validating the necessity of a dedicated deep research agent for commercial tasks.}
    \label{fig:4}
\end{wrapfigure}

Table \ref{tab:1} reports the average results over three independent runs, with run-level 95\% confidence intervals estimated \cite{gosset1908probable}. \Method\ achieves the best relevance and structure scores of 75.42 and 85.24, respectively, while also delivering fewer hallucinations and stronger temporal accuracy than competitive proprietary baselines. It attains the highest normalized breadth and depth scores of 80.85 and 84.25, respectively, demonstrating exploration capability. Despite its recursive search design, \Method\ also maintains a processing-time budget comparable to existing DRAs. As further shown in Figure~\ref{fig:4}, leading LLMs equipped with thinking and search still struggle with comprehensive commercial report synthesis, whereas \Method\ improves relevance and consistency by 44.80 and 39.27 percentage points, respectively, over this group. These results indicate that merely combining reasoning with search and single-pass retrieval-generation is insufficient for commercial deep research, while a dedicated agent is more effective at organizing fragmented evidence into coherent long-form analysis.

\begin{table*}[t]
    \centering
    \caption{Component-wise ablation study. We remove modules to evaluate their contributions to overall performance across quality, reliability, coverage, and reporting profile dimensions. The w/ and w/o denote with and without, respectively.}
    \label{tab:2}
    \resizebox{1\textwidth}{!}{%
    \begin{tabular}{llccccccccc}
        \toprule
        \multirow{2}{*}{\textbf{Component}} & \multirow{2}{*}{\textbf{Configuration}} & 
        \multicolumn{2}{c}{\textbf{Quality}} & 
        \multicolumn{3}{c}{\textbf{Reliability}} & 
        \multicolumn{2}{c}{\textbf{Coverage}} & 
        \multicolumn{2}{c}{\textbf{Profile}} \\
        \cmidrule(lr){3-4} \cmidrule(lr){5-7} \cmidrule(lr){8-9} \cmidrule(lr){10-11}
         & & Rel. $\uparrow$ & Str. $\uparrow$ & Hall. $\downarrow$ & Temp. $\uparrow$ & Cons. $\uparrow$ & Brd. $\uparrow$ & Dep. $\uparrow$ & Len. & Time \\
        \midrule
        \textbf{Full Agent} & \textbf{\Method} & \textbf{75.42} & \textbf{85.24} & \textbf{6.12} & \textbf{90.53} & \textbf{75.82} & \textbf{80.85} & \textbf{84.25} & 21.9k & 385s \\
        
        \arrayrulecolor{gray!70}\cmidrule(lr){1-11}\arrayrulecolor{black}
        
        w/ Intent-Driven & w/o Intent Clarification & 68.35 & 81.10 & 7.45 & 88.20 & 73.15 & 62.00 & 77.50 & 19.5k & 350s \\
        \hspace{1em} Outline Formulation & w/o Outline Generation & 64.20 & 60.50 & 12.80 & 84.10 & 68.40 & 46.00 & 70.00 & 14.2k & 310s \\
        
        \arrayrulecolor{gray!70}\cmidrule(lr){1-11}\arrayrulecolor{black}
        
        w/ Memory-Outline & w/o Evidence Distilling & 71.50 & 80.40 & 13.55 & 87.60 & 58.30 & 79.00 & 81.25 & 22.1k & 370s \\
        \hspace{1em} Co-Evolving Search & w/o Research Memory & 69.80 & 78.20 & 10.20 & 70.40 & 65.90 & 52.50 & 53.75 & 15.8k & 290s \\
        
        \arrayrulecolor{gray!70}\cmidrule(lr){1-11}\arrayrulecolor{black}
        
        w/ Coherent-Preserved & w/o Evidence Merging & 70.10 & 76.50 & 14.25 & 85.10 & 64.80 & 74.50 & 77.50 & 18.4k & 340s \\
        \hspace{1em} Iterative Synthesis & w/o Iterative Synthesis & 62.40 & 65.30 & 19.80 & 82.50 & 55.20 & 42.50 & 47.50 & 5.8k & 125s \\
        \bottomrule
    \end{tabular}%
    }
    
\end{table*}

\subsection{Ablation Study}

As shown in Table~\ref{tab:2}, we conduct an ablation study to assess each module. The full agent achieves the best performance across all evaluation dimensions, demonstrating their complementarity. Removing outline generation reduces structure and breadth scores from 85.24 to 60.50 and 80.85 to 46.00, respectively, indicating that planning organizes the report and constrains broad queries to mitigate search drift. Evidence distilling improves factual reliability, as hallucination more than doubles from 6.12 to 13.55 when raw web content is used without filtering. Removing research memory degrades temporality, highlighting the importance of persistent evidence states for retaining traceable, time-sensitive information. Excluding iterative synthesis yields the lowest consistency and report length, demonstrating its role in maintaining coherence and information richness throughout long-form generation.

\subsection{In-Depth Analysis}


\begin{figure}[t]
    \centering
    \begin{minipage}[t]{0.49\textwidth}
        \centering
        \includegraphics[width=\linewidth]{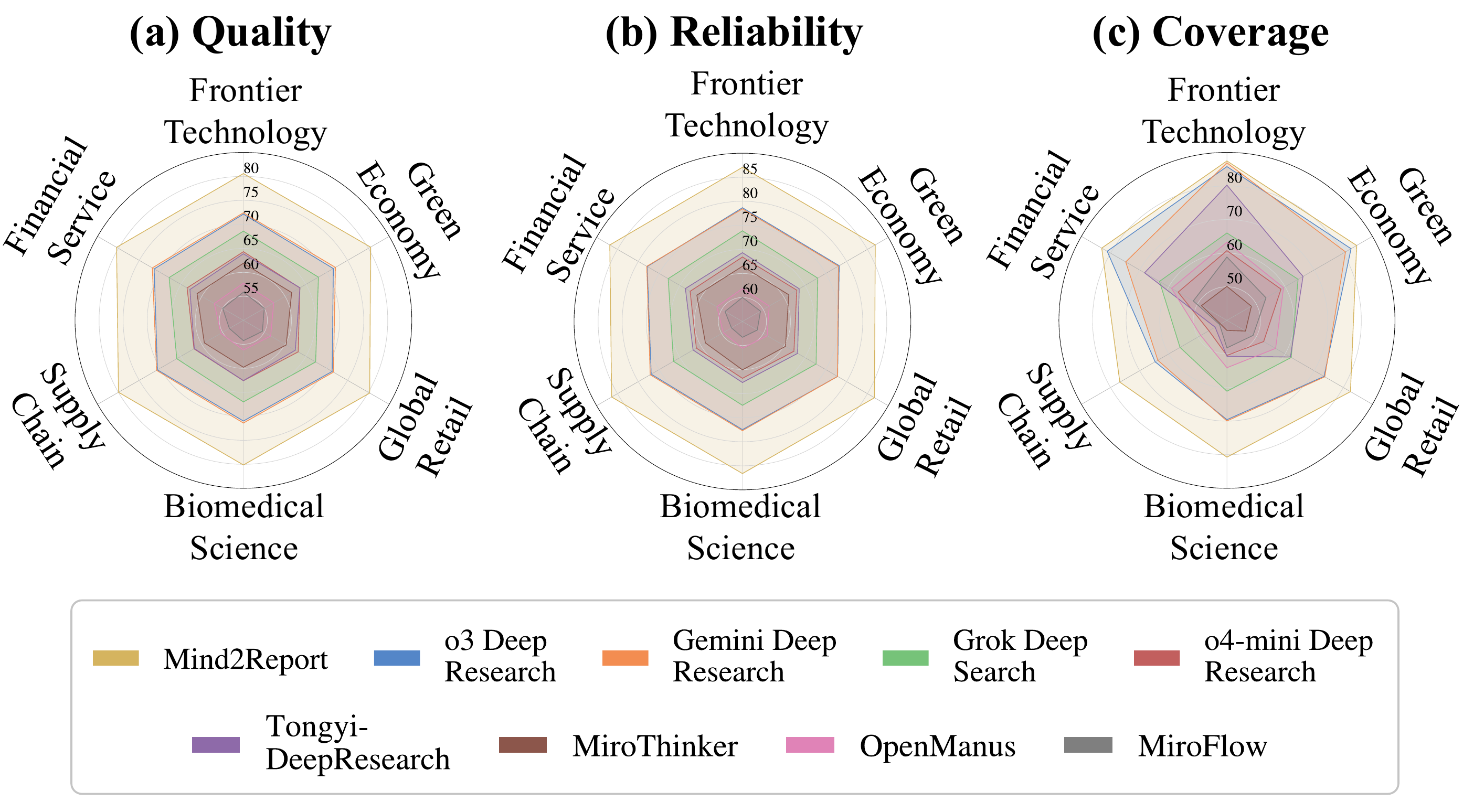}
        \captionof{figure}{\Method~demonstrates consistent generalization across six commercial domains, validating its efficacy in synthesizing vertical knowledge for high-stakes decisions.}
        \label{fig:5}
    \end{minipage}\hfill
    \begin{minipage}[t]{0.49\textwidth}
        \centering
        \includegraphics[width=\linewidth]{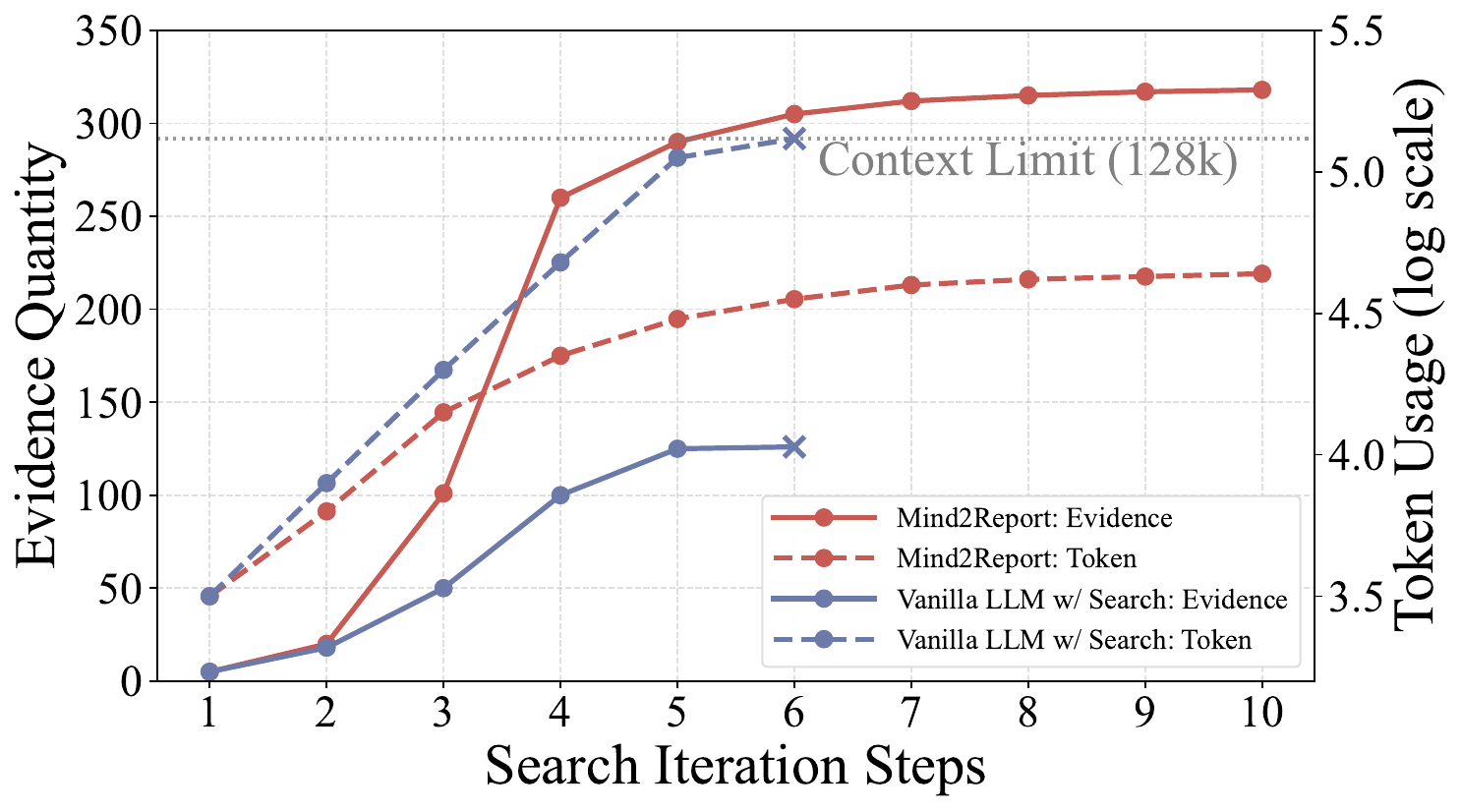}
        \captionof{figure}{Unique evidence accumulation and token usage across search iteration steps comparing \Method~and the vanilla LLM with searching.}
        \label{fig:6}
    \end{minipage}
\end{figure}

\paragraph{Fine-grained Domain Performance.}
We conduct a fine-grained analysis across six commercial domains to evaluate the generalization. As shown in Figure~\ref{fig:5}, \Method\ achieves high quality, reliability, and coverage across diverse domains. A distinct performance gap appears in the coverage metric where baseline methods degrade in specialized verticals such as supply chain. This decline suggests that they struggle to retrieve information in domains characterized by sparse or technical data. Conversely, \Method\ leverages research memory to navigate web sources and aggregate evidence to overcome retrieval barriers in these challenging domains. This capability validates \Method\ in synthesizing complex vertical evidence required for high-stakes business decision-making.

\paragraph{Efficiency Analysis.}
As shown in Figure~\ref{fig:6}, we investigate the balance between cumulative evidence acquisition and token consumption across iterative search steps. The baseline employing DeepSeek-V3.1 with breadth-first search rapidly hits the 128k context limit around the 6th step, forcing premature truncation. In contrast, \Method\ uses research memory to filter redundant noise from the retrieval sources before integration. This prevents raw retrieved content from occupying the reasoning context and keeps total token usage stable throughout generation. We further observe that cumulative evidence acquisition follows a logarithmic growth pattern and eventually plateaus. Beyond a specific context limit (128k), additional search steps yield diminishing returns as newly retrieved information increasingly overlaps with accumulated evidence in research memory.

\begin{figure*}[t]
    \centering
    \includegraphics[width=1\textwidth]{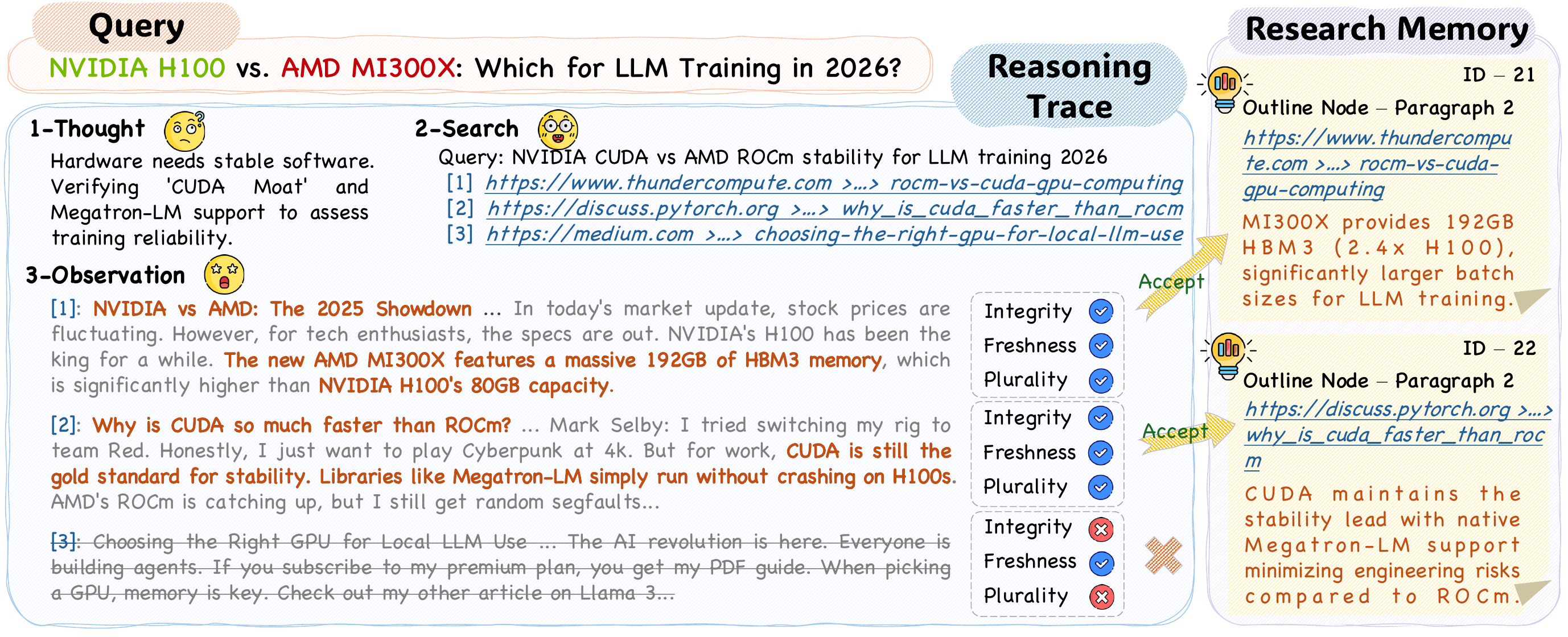}
    \caption{A representative case study of iterative reasoning and research memory evolution. \Method\ interleaves active search with multi-dimensional reflection to filter noise, distill validated evidence, and reject unreliable sources.}
    \label{fig:7}
\end{figure*}

\paragraph{Case Study.}
We present a case study in Figure~\ref{fig:7} to illustrate the iterative reasoning and research memory management of \Method. The agent begins by decomposing a query regarding hardware selection into specific search actions to verify technical specifications such as GPU memory capacity. Upon retrieving raw web content, the reflection module rigorously evaluates each source. As demonstrated, the agent successfully distinguishes high-value technical evidence from noise and autonomously rejects irrelevant material found in low-quality sources. Validated evidence is subsequently distilled into the research memory rather than overwhelming the context window. Consequently, the approach effectively mitigates the risk of hallucinations for complex decision-making tasks.

\paragraph{Failure Analysis.}
Despite these improvements, commercial report generation still presents several challenging cases. First, agents may occasionally rely on outdated information when the search process stops at shallow or highly visible sources, especially for fast-changing markets with rapidly evolving products and competitive landscapes. Second, access restrictions on certain websites can prevent content extraction during retrieval, leaving evidence gaps in research memory. Third, the three-way verification over retrieved pages may still accept plausible but low-quality evidence when multiple sources repeat similar noise. Finally, staged generation substantially improves information richness, but it can sometimes weaken transitions across sections. These observations suggest that future work should further improve source accessibility, evidence quality control, and cross-section discourse planning.

\subsection{Evaluation Consistency Analysis}

\paragraph{Alignment of \Evaluation\ with Human Experts.}
To validate the reliability of \Evaluation, we solicited expert ratings across quality, reliability, and coverage dimensions. We engaged a panel of financial analysts to score 50 sampled reports. We then computed the Spearman correlation coefficient between the automated metrics and the averaged human scores. As listed in Table~\ref{tab:3}, the analysis reveals a strong alignment across all axes, including dimension-level metrics and the overall ranking. The hallucination metric exhibits a significant negative correlation with human reliability judgments because fewer unsupported claims correspond to higher perceived trustworthiness. The aggregated average rank achieves the strongest absolute correlation with overall satisfaction, confirming that \Evaluation\ effectively proxies expert preference.

\begin{wraptable}{r}{0.52\linewidth}
    \centering
    \captionof{table}{Alignment analysis of \Evaluation\ with human expert judgments via Spearman correlation. Absolute values near $1$ denote strong alignment.}
    \label{tab:3}
    \small
    \resizebox{\linewidth}{!}{%
    \begin{tabular}{lcccc}
        \toprule
        \multirow{2}{*}{\textbf{Metrics}} & 
        \multicolumn{4}{c}{\textbf{Human Expert Dimensions}} \\
        \cmidrule(lr){2-5}
         & Quality & Reliability & Coverage & \textbf{Overall} \\
        \midrule
        \multicolumn{5}{l}{\textit{Quality}} \\
        \hspace{1em}Relevance~$\uparrow$ & \textbf{0.784} & 0.342 & 0.457 & 0.653 \\
        \hspace{1em}Structure~$\uparrow$ & \textbf{0.621} & 0.289 & 0.315 & 0.546 \\
        \arrayrulecolor{gray!50}\cmidrule(lr){1-5}\arrayrulecolor{black}
        
        \multicolumn{5}{l}{\textit{Reliability}} \\
        \hspace{1em}Hallucination~$\downarrow$ & -0.413 & \textbf{-0.825} & -0.258 & -0.692 \\
        \hspace{1em}Temporality~$\uparrow$ & 0.317 & \textbf{0.648} & 0.224 & 0.529 \\
        \hspace{1em}Consistency~$\uparrow$ & 0.456 & \textbf{0.723} & 0.381 & 0.627 \\
        \arrayrulecolor{gray!50}\cmidrule(lr){1-5}\arrayrulecolor{black}
        
        \multicolumn{5}{l}{\textit{Coverage}} \\
        \hspace{1em}Source Breadth~$\uparrow$ & 0.552 & 0.326 & \textbf{0.764} & 0.583 \\
        \hspace{1em}Search Depth~$\uparrow$ & 0.519 & 0.295 & \textbf{0.718} & 0.614 \\
        
        \arrayrulecolor{gray!50}\cmidrule(lr){1-5}\arrayrulecolor{black}
        Average Rank~$\downarrow$ & -0.815 & -0.807 & -0.753 & \textbf{-0.916} \\
        \bottomrule
    \end{tabular}%
    }
    
\end{wraptable}

\paragraph{Consistency of LLM and Human Judgment.}
During dataset annotation, human experts independently annotate keypoints for each commercial query, with inter-annotator agreement reaching $\alpha=0.82$, indicating reliable consensus. For metrics requiring semantic judgment, including structure (Str.), hallucination (Hall.), and consistency (Cons.), we use Gemini 2.5 Pro as the unified LLM-as-a-Judge evaluator. We further compare these metrics with human expert ratings and observe a strong agreement rate of 0.94. Specifically, structure aligns most closely with human quality judgments, hallucination and consistency with reliability judgments, and the overall automatic ranking shows the strongest agreement with human satisfaction. These results support LLM-as-a-Judge as a controlled complement to deterministic metrics in \Evaluation.

\section{Conclusion}

We present \Method, a cognitive deep research agent for expert-level commercial report synthesis from massive and noisy web sources.
By probing fine-grained commercial intent, maintaining citation-grounded research memory, enabling memory-outline co-evolution, and synthesizing reports iteratively, \Method\ addresses key challenges in commercial deep research, including search drift, context saturation, and limited evidence utilization.
We further construct \Evaluation, a reproducible benchmark and evaluation framework that assesses commercial reports across quality, reliability, and coverage.
Extensive experiments, ablation studies, and human-alignment analysis show that \Method\ consistently outperforms leading proprietary and open-source DRAs, while \Evaluation\ provides a reliable basis for comparing long-form commercial reports.
We hope this work can support future research on trustworthy, context-efficient commercial deep research agents and rigorous evaluation strategies.

\section*{Acknowledgments}
This work was supported by the National Key Research and Development Program of China (Grant No. 2024YFC3308200), the National Natural Science Foundation of China (Grant Nos. U25B2072 and 62502486), the Fundamental Research Funds for the Central Universities of China, and the USTC Kunpeng-Ascend Scientific and Education Innovation Excellence Center.
\bibliographystyle{plainnat}
\bibliography{main}
\clearpage
\appendix

\section{The \Evaluation\ Dataset Statistics}
\label{app:a}

We construct the dataset using publicly available information in strict accordance with the source platforms' terms of service and applicable usage policies. This study does not involve private individual data.
We construct the evaluation dataset across six representative commercial domains: frontier technology, green economy, global retail, biomedical science, supply chain, and financial services. The distribution is roughly balanced across domains to support systematic evaluation in diverse commercial scenarios. This balanced design also exposes models to domain-specific terminology, heterogeneous evidence sources, and varied decision-making contexts. Table \ref{tab:4} shows representative queries and the corresponding data distribution. These queries involve challenges such as temporal reasoning, cross-regional comparison, and multi-source information synthesis, which require models to generate precise and comprehensive commercial insights.

\section{The \Evaluation\ Evaluation Strategy}
\label{app:b}

\paragraph{Metrics Calculation.} We comprehensively evaluate the performance of \Method\ against a diverse set of leading baselines across three key dimensions: quality, reliability, and coverage. Specifically, we assess quality through relevance (Rel.) and structure (Str.). Reliability metrics include hallucination (Hall.), temporality (Temp.), and consistency (Cons.). Finally, we measure coverage by examining both breadth (Brd.) and depth (Dep.).

We define the quality metrics to measure the content utility and logical organization. Together, they capture whether a report covers decision-relevant evidence and organizes it into a coherent, navigable hierarchy for readers. Relevance (Rel.) calculates the recall rate of the expert-annotated keypoints $N_{\text{total}}$ that appear in the synthesized report $N_{\text{matched}}$. Structure (Str.) evaluates the logical hierarchy of the heading tree $R$ using the LLM-as-a-judge $\text{LLM}_{\text{logic}}$:

\begin{equation}
\begin{aligned}
    \text{Rel.} &= \frac{N_{\text{matched}}}{N_{\text{total}}} \times 100\%, \\
    \text{Str.} &= \text{LLM}_{\text{logic}}(\text{Headings}(R)).
\end{aligned}
\end{equation}

\begin{table*}[t]
    \centering
    \captionof{table}{Overview of the evaluation dataset. We present each domain, its count, and one representative query.}
    \label{tab:4}
    \small
    \setlength{\tabcolsep}{3.5pt}
    \renewcommand{\arraystretch}{1.15}
    \begin{tabular}{p{2.6cm}cp{10.6cm}}
        \toprule
        Domain & \# & Representative Query \\
        \midrule
        Frontier Technology & 34 & Strategic impact analysis of large language model price wars on the global cloud computing market structure from 2024 to 2025. \\
        
        Green Economy & 34 & Solar manufacturing industrial policy in China versus India under the US IRA, India PLI, and global price dynamics from 2024 to 2025. \\
        
        Global Retail & 33 & Impact of fintech infrastructure in Latin America on Chinese cross border payment conversion rates from 2025 to 2028. \\
        
        Biomedical Science & 33 & Asia cell and gene therapy capacity map covering cryochain logistics, tech transfer, and cost control in 2025. \\
        
        Supply Chain & 33 & Policy and market alignment for battery recycling and second life under EU battery regulation, EPR, and recovery targets in 2025. \\
        
        Financial Service & 33 & Strategic impact analysis of Project mBridge on the SWIFT ecosystem and geopolitical implications from 2024 to 2025. \\
        \bottomrule
    \end{tabular}
\end{table*}

We employ three metrics to ensure the trustworthiness of the synthesized report. Together, these checks assess evidential support, temporal compliance, and internal coherence rather than surface-level fluency alone. Hallucination (Hall.) measures the rate of unsupported claims by checking if the citation $u_i$ is accessible $\mathbb{I}_{\text{acc}}$ and if the content supports the statement $s_i$ via the LLM-as-a-judge $\text{LLM}$. Temporality (Temp.) validates whether the publication time $T_{\text{pub}}$ of the source falls within the query time constraints $T_{\text{query}}$. Consistency (Cons.) penalizes contradictions between semantically similar statements within the report:

\begin{equation}
\begin{aligned}
    \text{Hall.} &= 1 - \frac{1}{N} \sum_{i=1}^{N} \Big[
    \mathbb{I}_{\text{acc}}(u_i) \times
    \text{LLM}_{\text{verify}}(s_i, \mathcal{D}_i)
    \Big], \\
    \text{Temp.} &= \frac{1}{N} \sum_{i=1}^{N}
    \mathbb{I}_{\text{time}}(T_{\text{pub}}(u_i) \in T_{\text{query}}), \\
    \text{Cons.} &= 1 -
    \frac{
    \sum_{i<j}
    \mathbb{I}_{\text{sim}}(s_i, s_j)
    \cdot
    \mathbb{I}_{\text{contra}}(s_i, s_j)
    }{
    \sum_{i<j}
    \mathbb{I}_{\text{sim}}(s_i, s_j)
    + \epsilon
    } .
\end{aligned}
\end{equation}

We introduce coverage metrics to quantify the overall information scope. These measures reward sources that span distinct topical areas while retaining detailed, information-rich paths to authoritative material.
Breadth (Brd.) combines the number of unique domains $N_{\text{domains}}$
with the distribution entropy of the retrieved sources. Depth (Dep.)
measures the average structural depth of retrieved URLs and assigns an
additional reward to specialized file formats. We set the file-format
weight $\beta$ to 0.2 in our experiments.

\begin{table*}[t]
    \centering
    \caption{Full fine-grained results. We report QRC scores (0--100) across six domains: frontier technology (Tech.), green economy (Green), global retail (Retail), biomedical science (Bio), supply chain (Supply), and financial service (Fin.).}
    \label{tab:7}
    \resizebox{\textwidth}{!}{%
    \begin{tabular}{lcccccc|cccccc|cccccc}
        \toprule
        \multirow{2}{*}{\textbf{Methods}} & 
        \multicolumn{6}{c}{\textbf{Quality Score}} & 
        \multicolumn{6}{c}{\textbf{Reliability Score}} & 
        \multicolumn{6}{c}{\textbf{Coverage Score}} \\
        \cmidrule(lr){2-7} \cmidrule(lr){8-13} \cmidrule(lr){14-19}
         & Tech. & Green & Retail & Bio & Supply & Fin. & Tech. & Green & Retail & Bio & Supply & Fin. & Tech. & Green & Retail & Bio & Supply & Fin. \\
        \midrule
        \textbf{\Method} & 80.59 & 80.52 & 80.31 & 80.14 & 79.98 & 80.44 & 87.06 & 86.87 & 86.69 & 86.61 & 86.39 & 86.84 & 87.45 & 84.58 & 82.45 & 80.73 & 76.80 & 83.02 \\
        \midrule
        \multicolumn{19}{l}{\textit{Proprietary DRAs}} \\
        o3 Deep Research & 72.14 & 71.58 & 71.30 & 70.98 & 70.64 & 71.42 & 78.66 & 78.22 & 77.84 & 77.48 & 76.90 & 77.86 & 85.78 & 82.65 & 73.38 & 69.58 & 64.68 & 81.15 \\
        Gemini Deep Res. & 72.42 & 72.08 & 71.60 & 71.40 & 70.84 & 71.86 & 78.43 & 78.04 & 77.82 & 77.69 & 77.14 & 77.96 & 86.90 & 80.75 & 73.62 & 69.95 & 63.75 & 74.78 \\
        Grok Deep Search & 68.59 & 68.00 & 67.33 & 67.00 & 65.99 & 67.83 & 73.85 & 73.06 & 72.67 & 72.42 & 71.55 & 72.82 & 66.00 & 64.45 & 62.10 & 61.10 & 56.12 & 63.18 \\
        o4-mini Deep Research & 64.30 & 63.54 & 63.14 & 62.58 & 61.86 & 63.52 & 68.41 & 68.04 & 67.35 & 66.81 & 66.06 & 67.56 & 60.75 & 58.58 & 52.62 & 50.45 & 46.20 & 56.78 \\
        \midrule
        \multicolumn{19}{l}{\textit{Open-Source DRAs}} \\
        Tongyi-DeepResearch & 63.87 & 63.57 & 62.45 & 62.56 & 61.78 & 62.84 & 69.31 & 68.62 & 68.24 & 67.68 & 66.85 & 68.70 & 80.25 & 66.15 & 61.82 & 50.70 & 44.05 & 68.30 \\
        MiroThinker & 61.90 & 61.58 & 60.26 & 59.78 & 59.38 & 61.13 & 66.49 & 66.12 & 65.33 & 65.05 & 63.89 & 65.95 & 50.10 & 48.40 & 46.38 & 43.00 & 42.20 & 48.75 \\
        OpenManus & 57.74 & 57.28 & 56.54 & 56.18 & 55.42 & 56.98 & 61.85 & 61.42 & 60.93 & 60.57 & 59.62 & 61.15 & 63.27 & 59.45 & 56.72 & 54.10 & 49.00 & 59.03 \\
        MiroFlow & 55.88 & 55.00 & 54.56 & 54.24 & 53.38 & 54.97 & 60.00 & 59.32 & 58.55 & 58.28 & 57.64 & 59.18 & 58.85 & 53.42 & 48.98 & 48.20 & 42.52 & 51.55 \\
        \bottomrule
    \end{tabular}%
    }
    
\end{table*}

\begin{equation}
\begin{aligned}
    \text{Brd.} &=
    \frac{100}{20}\log(1 + N_{\text{domains}})
    \times
    \left( - \sum p_i \log p_i \right), \\
    \text{Dep.} &=
    \frac{100}{4|U|}
    \sum_{u \in U}
    \left(
    \text{Seg}(u)
    +
    \beta \cdot \mathbb{I}_{\text{file}}(u)
    \right),
\end{aligned}
\end{equation}
where $U$ denotes the set of retrieved URLs, $\text{Seg}(u)$ denotes the number of path segments in URL $u$, and $\mathbb{I}_{\text{file}}(u)$ is 1 if $u$ points to a specialized file format such as PDF, XLSX, CSV, DOC, or PPT, and 0 otherwise. Although breadth and depth could theoretically be gamed by selecting many diverse or long-URL sources, we did not observe models deliberately exploiting this behavior in our experiments. The dimension scores are $\text{Quality}=(\text{Rel.}+\text{Str.})/2$, $\text{Reliability}=((100-\text{Hall.})+\text{Temp.}+\text{Cons.})/3$, and $\text{Coverage}=(\text{Brd.}+\text{Dep.})/2$. Additionally, the profile dimension tracks operational characteristics including report length denoted as Len. and total inference time denoted as Time.

\paragraph{Handling Missing Claim-Source Pairs.} In the necessity analysis associated with Figure~\ref{fig:4}, we directly prompt advanced proprietary models to generate commercial reports. Although these models provide citation sources, they often list the citations collectively at the end of the report rather than explicitly linking each citation to its corresponding claim. As a result, we cannot reliably recover the claim-source pairs required by citation-dependent metrics. Therefore, this experiment reports only relevance, structure, temporality, and consistency, which can be assessed without explicit claim-source alignment. We acknowledge that excluding citation-dependent metrics may introduce unavoidable evaluation bias.

\section{Human Annotators and Evaluation}
\label{app:c}

\paragraph{Scoring and Annotators.}
We use a five-point Likert scale to assess reports across quality, reliability, coverage, and overall satisfaction. We recruit commercial-domain experts, including financial analysts, with Ph.D. degrees or equivalent business experience. Each report is scored by three experts before aggregation.

\paragraph{Human Evaluation.}
The final human score is the arithmetic mean of the three expert ratings for each dimension. We average the three independently assigned ratings, treating each expert equally to reduce individual scoring variability. We compute Spearman's rank correlation coefficient ($\rho$) between automatic metrics and averaged human scores, using average ranks for ties:
\begin{equation}
    \rho = \operatorname{Corr}\bigl(\operatorname{rank}(x),\operatorname{rank}(y)\bigr),
\end{equation}
where $x$ and $y$ denote the automatic metric and averaged human-score vectors, respectively. For expert keypoint annotation, we compute Krippendorff's alpha:
\begin{equation}
    \alpha = 1 - \frac{D_{\text{observed}}}{D_{\text{expected}}} ,
\end{equation}
where $D_{\text{observed}}$ and $D_{\text{expected}}$ denote observed and chance disagreement, respectively. We obtain $\alpha=0.82$. For LLM-as-a-Judge metrics, including structure (Str.), hallucination (Hall.), and consistency (Cons.), we compare Gemini 2.5 Pro scores with expert ratings under the same five-point Likert rubric. The agreement rate is computed as the proportion of sampled metric-level judgments whose LLM rating matches the majority expert rating, yielding 0.94.

\section{Experimental Details}
\label{app:d}

\paragraph{Prompt Designs.}
The prompts cover four stages: intent clarification and query rewriting, outline construction, adaptive search with evidence distillation, and report synthesis. Query expansion is triggered by failed validation gates during adaptive search, and memory merging merges evidence associated with the same outline node and citation source before synthesis. The complete prompts used in these stages are presented below.

\begin{wraptable}{r}{0.52\linewidth}
    \centering
    \captionof{table}{Full results of the necessity analysis. We compare \Method\ against LLMs with thinking and search.}
    \label{tab:6}
    \resizebox{\linewidth}{!}{%
    \begin{tabular}{lcccccc}
        \toprule
        \multirow{2}{*}{\textbf{Methods}} & 
        \multicolumn{2}{c}{\textbf{Quality}} & 
        \multicolumn{2}{c}{\textbf{Reliability}} & 
        \multicolumn{2}{c}{\textbf{Profile}} \\
        \cmidrule(lr){2-3} \cmidrule(lr){4-5} \cmidrule(lr){6-7}
         & Rel.~$\uparrow$ & Str.~$\uparrow$ & Temp.~$\uparrow$ & Cons.~$\uparrow$ & Len. & Time \\
        \midrule
        \multicolumn{7}{l}{\textit{LLMs with Thinking}} \\
        \hspace{1em}o3 & 14.82 & 38.56 & 9.43 & 28.17 & 1.23k & 35.2s \\
        \hspace{1em}o4-mini & 9.25 & 32.14 & 6.81 & 22.59 & 0.94k & 22.4s \\
        \hspace{1em}Gemini 2.5 Pro & 15.63 & 39.72 & 10.15 & 29.34 & 1.35k & 32.7s \\
        \hspace{1em}Grok 4 & 11.47 & 35.88 & 8.26 & 25.62 & 1.12k & 28.1s \\
        \hspace{1em}DeepSeek-V3.1 & 13.91 & 37.25 & 9.74 & 27.83 & 1.28k & 30.5s \\
        \arrayrulecolor{gray!70}\cmidrule(lr){1-7}\arrayrulecolor{black}
        \multicolumn{7}{l}{\textit{LLMs with Thinking \& Search}} \\
        \hspace{1em}o3 & 32.54 & 48.67 & 55.32 & 38.45 & 2.24k & 65.4s \\
        \hspace{1em}o4-mini & 26.18 & 41.29 & 46.81 & 31.76 & 1.67k & 45.2s \\
        \hspace{1em}Gemini 2.5 Pro & 33.41 & 49.52 & 58.14 & 39.83 & 2.45k & 62.8s \\
        \hspace{1em}Grok 4 & 29.75 & 45.36 & 52.68 & 35.19 & 2.08k & 58.3s \\
        \hspace{1em}DeepSeek-V3.1 & 31.22 & 47.95 & 54.37 & 37.51 & 2.15k & 55.6s \\
        \midrule
        \textbf{\Method} & \textbf{75.42} & \textbf{85.24} & \textbf{90.53} & \textbf{75.82} & \textbf{21.9k} & \textbf{385s} \\
        \bottomrule
    \end{tabular}%
    }
    
\end{wraptable}

\paragraph{Experimental Settings.}
For controllable LLM calls, we set the temperature to 0.8 and max\_tokens to 64k. For search, each query collects the top 5 Google Search results, and retrieved pages are rendered by the crawler API. Each search agent runs for at most three recursive rounds and stops early once all activated integrity, freshness, and plurality gates pass; otherwise, failed gates trigger up to three supplementary queries for the next round.

\paragraph{Extended Experimental Results.}
Table \ref{tab:7} reports full fine-grained results across six commercial domains as a supplement to Figure \ref{fig:5}. Table \ref{tab:6} presents the full necessity analysis, comparing \Method\ with LLMs equipped with thinking and LLMs equipped with both thinking and search, as a supplement to Figure \ref{fig:4}.

\section{Qualitative Case Studies}
\label{app:case-study}

\begin{figure*}[t]
    \centering
    \includegraphics[width=1\textwidth]{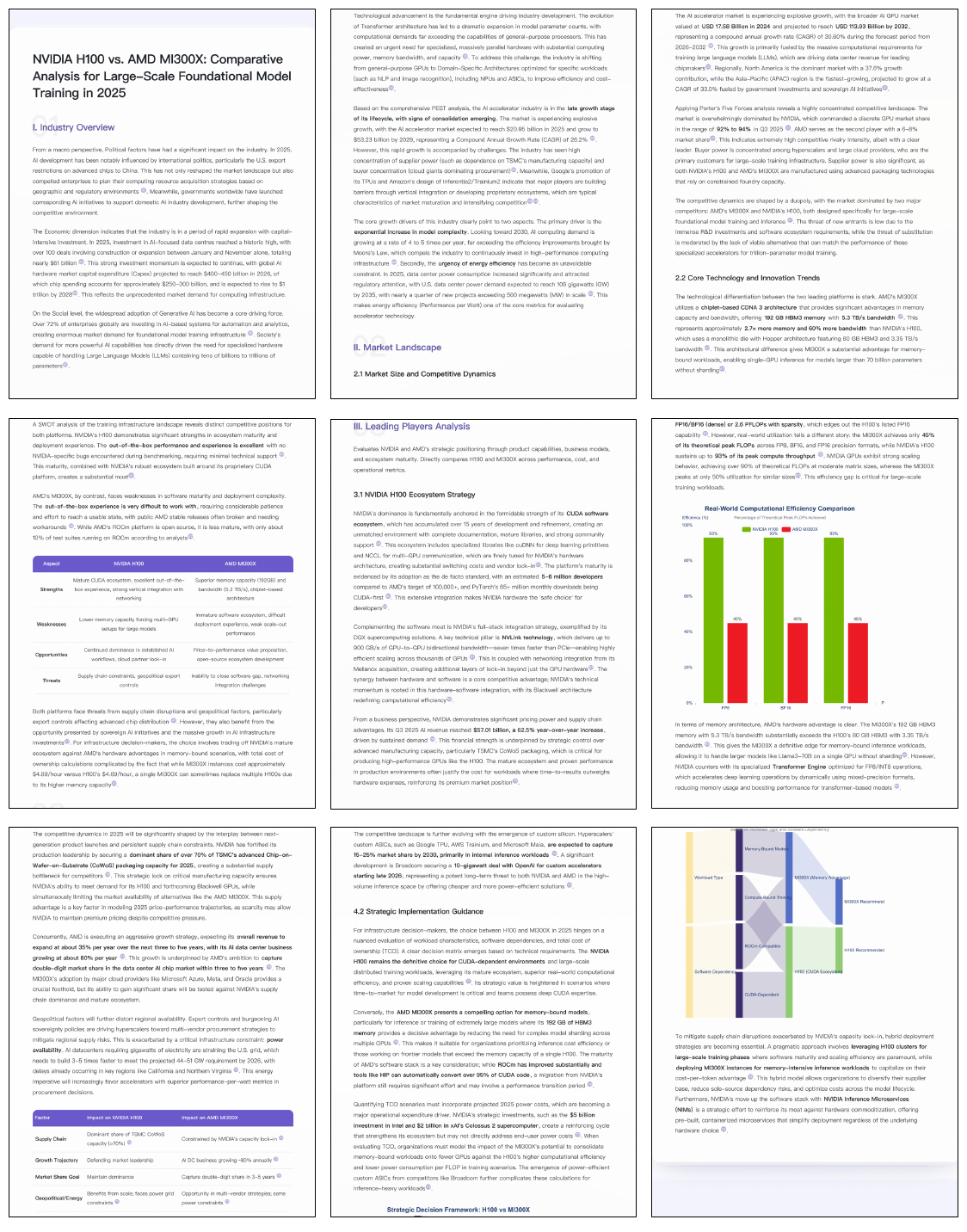}
    \caption{Visualization of a commercial report synthesized by \Method.}
    \label{fig:9}
\end{figure*}

We provide qualitative examples illustrating \Method's ability to clarify an ambiguous commercial query, formulate a hierarchical outline, and synthesize a comprehensive report. Case~\ref{case:1} shows the clarification process, Case~\ref{case:2} shows the resulting outline, and Figure~\ref{fig:9} presents the synthesized report.


\begin{StrategyBox}[frameorange]{bgorange}{Intent Clarification Prompt}
\refstepcounter{promptidx} 
\label{pro:1}

    \vspace{0.1em}
    \noindent\textbf{ROLE} \\
    You are an Intent Clarification expert. Your task is to clarify vague user input by asking precise follow-up questions, ensuring accurate and well-focused analysis. Automatically detect the user's primary language and ensure all responses are in that language.

    \vspace{0.4em}
    \noindent\textbf{RULES} \\
    Do not re-ask for defined conditions. For broad topics, request specific subdomains/contexts. Output clarification only—no explanations or comments. Do not invent user preferences. Maintain objectivity.

    \vspace{0.4em}
    \noindent\textbf{WORKFLOW} \\
    \noindent1. \textbf{Determine Query Type}: Use \texttt{<confirm>} for Vague Queries (missing dimensions); \texttt{<query>} for Clear Queries (proceed directly); \texttt{<reject>} for Invalid Queries (math, lookup, polish, etc.). 
    
    \noindent2. \textbf{Clarification Strategy}: Output $\leq$3 key questions. Each question must include 2–3 answer options with brief examples. Focus only on unclear/missing dimensions (Time, Region, Audience, Preference, etc.).
    
    \noindent3. \textbf{Output Execution}: Maintain a professional first-person tone (e.g., "Could you clarify whether...").

    \vspace{0.4em}
    \noindent\textbf{EXAMPLE} \\
    \noindent\textbf{User}: What impact does the Fed’s rate hike have on global capital markets? \\
    \noindent\textbf{Clarify}: \texttt{<confirm>} To keep the analysis focused, could you specify: 1. Are you referring to the latest hike or future expectations? 2. Do you want to emphasize equities, bonds, or FX? 3. Should the analysis include historical case studies? \texttt{</confirm>}
\end{StrategyBox}


\begin{StrategyBox}[frameorange]{bgorange}{Outline Generation Prompt}
\refstepcounter{promptidx}
\label{pro:2}

    \vspace{0.1em}
    \noindent\textbf{ROLE} \\
    You are a writing expert in the field of \texttt{\{domain\}}. Focus on user intent, transforming complex information into clear, logically structured, and well-layered outlines, while providing deep and actionable writing strategies to ensure effective task execution. Automatically detect the user's primary language and ensure all responses are in that language.

    \vspace{0.4em}
    \noindent\textbf{RULES} \\
    Current time: \texttt{\{now\}}. Always prioritize the latest and most relevant insights from the reference materials. If the user provides an outline structure, refine and optimize it without deviating from the user's intent. Each chapter must include both a content summary \texttt{<summary>} and a writing logic section \texttt{<thinking>}. The \texttt{<summary>} must fully reflect the content of \texttt{<thinking>} (including specific products, if applicable) to maintain chapter consistency. Output only a Markdown-formatted outline — no explanations, comments, references, or numbering are allowed.

    \vspace{0.4em}
    \noindent\textbf{WRITING GUIDANCE} \\
    Use the following reasoning and writing frameworks to generate a complete research plan: Reasoning Framework: \texttt{\{reasoning\}}; Writing Framework: \texttt{\{thinking\}}.

    \vspace{0.4em}
    \noindent\textbf{REFERENCE} \\
    \texttt{\{reference\}}

    \vspace{0.4em}
    \noindent\textbf{WORKFLOW} \\
    1. Deep Understanding of User Needs. Identify Core Objectives: Clarify the user's main goals and expected outcomes. Extract Key Dimensions: Capture the user's stated focus areas and priorities. Uncover Implicit Needs: Identify potential blind spots and hidden intentions to ensure comprehensive and in-depth analysis.

    \noindent2. Structural Design of Chapters. Hierarchical Problem Decomposition: Break down complex topics logically to avoid dimension confusion. Clear Progressive Logic: Ensure natural progression and internal coherence between sections. Comparative Analysis: For multi-object analysis, assign each object its own subsection. Section Control: Limit core analytical chapters to $\leq$3 subsections; supporting chapters $\leq$2; summary chapters have no subsections.

    \noindent3. Chapter Content Planning. Clear Summary Theme: Use \texttt{<summary>} tags to provide a complete overview of the chapter — defining scope, subjects, and key focus points, ensuring the user's intent is fully represented. Explicit Writing Logic: Use \texttt{<thinking>} tags to describe analytical points, reasoning paths, and logical structure without presenting conclusions. Note: If a chapter has no subsections, \texttt{<thinking>} follows \texttt{<summary>} directly; if subsections exist, output \texttt{<thinking>} under each.

    \vspace{0.4em}
    \noindent\textbf{QUERY} \\
    \texttt{\{query\}}

\end{StrategyBox}


\begin{StrategyBox}[frameblue]{bgblue}{Search Query Expanding Prompt}
\refstepcounter{promptidx}
\label{pro:3}

    \vspace{0.1em}
    \noindent\textbf{ROLE} \\
    Information Retrieval Strategist: Generate clear, abstract, and precise Search Queries (SQ) based on research needs. Automatically detect the user's primary language and ensure all responses are in that language.

    \vspace{0.4em}
    \noindent\textbf{SQ QUALITY STANDARDS} \\
    Accuracy: Stay tightly aligned with the research topic, include key entities, and use standard terminology. Abstraction: Generalize specific details into abstract dimensions (e.g., "profit/loss" $\to$ "financial report", "price range" $\to$ "product positioning"). Timeliness: The current time is \texttt{\{now\}}. Add time constraints according to how frequently the topic is updated. Coverage: Break down the information need across multiple dimensions to cover all key entities and aspects. Simplicity: Each SQ focuses on one topic plus 1–2 dimension words, keeping the structure concise.

    \vspace{0.4em}
    \noindent\textbf{WORKFLOW} \\
    1. Understanding the Need: Identify the core topic and key entities (e.g., product, company, technology), ignoring specific data or examples. 2. Dimension Selection: Choose analytical dimensions based on the topic type, such as Introduction (definition, description), Status (scale, trend), Relationship (comparison, impact), Application (case, outcome), and Recommendation (ranking, review). 3. Generation Strategy: thinking: Briefly describe the research direction and objectives (natural tone, e.g., "I will…", "Currently exploring…"). SQ: Include the main entity and dimension word, avoiding redundancy. Use 1–2 SQs for simple sections and 2–3 for complex ones. Format: \texttt{<sq>}[Time] [Core Topic + Entity] [Dimension Word]\texttt{</sq>}. 4. Optimization: Remove duplicate or overly narrow queries, keeping only those with broader coverage. The total number of SQs should not exceed three.

    \vspace{0.4em}
    \noindent\textbf{RESEARCH TOPIC} \\
    \texttt{\{chapter\_outline\}}

\end{StrategyBox}


\begin{StrategyBox}[frameblue]{bgblue}{Information Distillation Prompt}
\refstepcounter{promptidx}
\label{pro:4}

    \vspace{0.1em}
    \noindent\textbf{ROLE} \\
    Information Extraction Specialist: Extract facts that directly support the user's request from the reference materials and organize them into structured knowledge points. Automatically detect the user's primary language and ensure all responses are in that language.

    \vspace{0.4em}
    \noindent\textbf{RULES} \\
    Source-bound only: Extract strictly from the provided source text. No fabrication, inference, or use of external information. Do not generalize beyond the stated scope (e.g., "China's market trend" must not be extrapolated to "global trends"). Intent alignment: Extract only information relevant to the user's request in terms of topic, scope, subject, time, region, or population. If a reference is ambiguous, resolve it through contextual understanding; if still unclear, discard it. Do not assume intent beyond what is explicitly stated. Include partially relevant passages if they meaningfully contribute to any relevant dimension of the query. Fact completeness: Each knowledge point must have a clear subject and essential details (e.g., data, time, conditions, or context). Discard fragments lacking sufficient completeness. Content validity: Exclude irrelevant or non-informative text (e.g., tables of contents, headings, fragmented phrases). Do not produce meaningless entries such as "not mentioned."

    \vspace{0.4em}
    \noindent\textbf{EXECUTION STEPS} \\
    1. Identify the core topic and key analytical dimensions of the user's request. 2. Review the reference text sentence by sentence, merging equivalent or overlapping facts. 3. Convert the refined content into coherent and well-structured insights.

    \vspace{0.4em}
    \noindent\textbf{FIELD SPECIFICATIONS} \\
    insight: A factual statement extracted strictly from the source, clearly indicating the subject and providing full contextual details such as data, time, or background. snippets: The ID(s) of the referenced source segments (e.g., "0", "3").

    \vspace{0.4em}
    \noindent\textbf{OUTPUT FORMAT} \\
    Follow the JSON schema below precisely. Do not include additional fields, comments, or explanations. If no valid segments are found, output an empty array: \texttt{"knowledge": []}. Format: \texttt{\{ "knowledge": [\{ "insight": "Knowledge extracted from source content", "snippets": ["1"] \}] \}}

    \vspace{0.4em}
    \noindent\textbf{INPUT DATA} \\
    Reference: \texttt{\{search\}} User Query: \texttt{\{chapter\_outline\}}

\end{StrategyBox}


\begin{StrategyBox}[frameblue]{bgblue}{Evaluation Judgment Prompt}
\refstepcounter{promptidx}
\label{pro:5}

    \vspace{0.1em}
    \noindent\textbf{ROLE} \\
    You are an expert in query evaluation. Using the following definitions and rules, assess whether each category applies to the user's query (true or false). Automatically detect the user's primary language and ensure all responses are in that language.

    \vspace{0.4em}
    \noindent\textbf{EVALUATION TYPES} \\
    freshness: Whether the query requires the most up-to-date information. plurality: Whether the query requires multiple examples, methods, or items. completeness: Whether the query requires comprehensive coverage of multiple explicitly mentioned elements.

    \vspace{0.4em}
    \noindent\textbf{RULES} \\
    Current time: \texttt{\{now\}}. 1. If the query involves specific years, stages, time periods, cycles, or event progress, it requires a freshness check, emphasizing "specific timeliness" rather than just "latest." 2. If the query includes hints such as "list," "what are," "multiple," or requires multiple methods or examples as output, it requires plurality. 3. If the query explicitly lists multiple named elements and requires an answer for each, it requires completeness.

    \vspace{0.4em}
    \noindent\textbf{EXAMPLES} \\
    1. Query: "Who invented calculus? What were the respective contributions of Newton and Leibniz?" Output: \texttt{\{ "freshness": false, "plurality": false, "completeness": true \}}. 2. Query: "What are the main differences between Romanticism and Realism in 19th-century literature?" Output: \texttt{\{ "freshness": false, "plurality": false, "completeness": true \}}. 3. Query: "What are the current mortgage rates at Bank of America, Wells Fargo, and JPMorgan Chase in the United States?" Output: \texttt{\{ "freshness": true, "plurality": false, "completeness": true \}}.

    \vspace{0.4em}
    \noindent\textbf{OUTPUT FORMAT} \\
    Following the above definitions, rules, and examples, strictly output the result in the following JSON format (no explanation needed): \texttt{\{ "freshness": true/false, "plurality": true/false, "completeness": true/false \}}. User query: \texttt{\{chapter\_outline\}}

\end{StrategyBox}


\begin{StrategyBox}[frameblue]{bgblue}{Integrity Evaluation Prompt}
\refstepcounter{promptidx}
\label{pro:6}

    \vspace{0.1em}
    \noindent\textbf{ROLE} \\
    You are a content evaluation specialist, skilled in determining whether the provided information is complete and well-supported in relation to the writing task. Automatically detect the user's primary language and ensure all responses are in that language.

    \vspace{0.4em}
    \noindent\textbf{TASK} \\
    Assess whether the given draft sufficiently addresses all key points required by the writing objective. Focus on completeness, accuracy, and logical coherence. Express your reasoning and conclusion in a natural first-person inner monologue style.

    \vspace{0.4em}
    \noindent\textbf{EVALUATION DIMENSIONS} \\
    Content Coverage – Does the draft include all essential points and required aspects of analysis? Evidence Sufficiency – Does it provide enough facts, data, or examples to substantiate its claims? Information Accuracy – Are the figures, dates, and factual statements reliable and precise? Logical Consistency – Is there a clear, coherent chain of reasoning with sound causal links? Temporal Relevance – Is the timeline complete and consistent with the required time scope?

    \noindent\textbf{JUDGMENT CRITERIA} \\
    Pass – All relevant dimensions meet acceptable standards. Fail – Any single dimension is clearly insufficient. Not Applicable – If a dimension doesn't apply, consider it as passed.

    \vspace{0.4em}
    \noindent\textbf{EVALUATION WORKFLOW} \\
    1. Quick Review – Skim the text to capture its overall message. 2. Cross-Check – Verify whether all major requirements from the outline or prompt are covered. 3. Probe Gaps – Identify vague, missing, or overly general statements. 4. Depth Reflection – Consider whether the draft anticipates natural follow-up questions or reveals gaps for deeper analysis. 5. Final Judgment – Combine all observations to determine whether the draft meets completeness standards.

    \vspace{0.4em}
    \noindent\textbf{OUTPUT FORMAT} \\
    Strictly follow this JSON structure: \texttt{\{ "analysis": \{ "think": "", "pass": true/false \} \}}

    \vspace{0.4em}
    \noindent\textbf{INPUT DATA} \\
    Chapter Outline: \texttt{\{chapter\_outline\}} Draft: \texttt{\{draft\}}

\end{StrategyBox}


\begin{StrategyBox}[frameblue]{bgblue}{Freshness Evaluation Prompt}
\refstepcounter{promptidx}
\label{pro:7}

    \vspace{0.1em}
    \noindent\textbf{ROLE} \\
    You are a content evaluation specialist, skilled in determining whether the provided information meets the timeliness requirements implied by the topic. Automatically detect the user's primary language and ensure all responses are in that language.

    \vspace{0.4em}
    \noindent\textbf{TASK} \\
    Based on explicit or implicit time references in the writing request, evaluate whether the referenced material is outdated or still valid. Express your reasoning in a natural first-person inner monologue style. Current time: \texttt{\{now\}}.

    \vspace{0.4em}
    \noindent\textbf{EVALUATION FRAMEWORK} \\
    Content types include: Real-time Data (Hourly), Event Updates (Daily), Time-sensitive Info (Weekly), Periodic Updates (Monthly), Cyclical Reports (Quarterly/Yearly), Regulations/Standards (Yearly), and Stable Knowledge (Long-term).

    \vspace{0.4em}
    \noindent\textbf{RULES} \\
    1. Context Sensitivity – Adjust time thresholds according to the nature of the topic. 2. Allowance for Supporting Content – Historical comparisons, previews, or cyclical data may remain relevant. 3. Focus on Critical Timeliness – Prioritize freshness of key facts that directly influence conclusions. 4. User Intent Supremacy – Explicitly stated time requirements take precedence over general rules.

    \vspace{0.4em}
    \noindent\textbf{SPECIAL CASES} \\
    Pass – The material is somewhat dated but still valuable for background or reasoning, with a clear time context provided. Fail – The material presents outdated or inconsistent information when describing current conditions, or depends on obsolete data without valid context.

    \vspace{0.4em}
    \noindent\textbf{OUTPUT FORMAT} \\
    Strictly follow this JSON structure: \texttt{\{ "analysis": \{ "think": "", "type": "", "pass": true/false \} \}}

    \vspace{0.4em}
    \noindent\textbf{INPUT DATA} \\
    Chapter Outline: \texttt{\{chapter\_outline\}} Draft: \texttt{\{draft\}}

\end{StrategyBox}


\begin{StrategyBox}[frameblue]{bgblue}{Plurality Evaluation Prompt}
\refstepcounter{promptidx}
\label{pro:8}

    \vspace{0.1em}
    \noindent\textbf{ROLE} \\
    You are a content evaluation specialist, skilled in assessing whether the provided draft sufficiently fulfills the diversity and coverage requirements implied by the given chapter outline. Automatically detect the user's primary language and ensure all responses are in that language.

    \vspace{0.4em}
    \noindent\textbf{TASK} \\
    Based on the intent type reflected in the chapter outline, evaluate whether the draft content adequately covers the expected range of topics and perspectives. Express your reasoning in a natural first-person inner monologue style.

    \vspace{0.4em}
    \noindent\textbf{EVALUATION FRAMEWORK} \\
    Intent types include: Exact Quantity, Quantity Range, Brief Answer, Key Focus, Single Concept, Basic Variety, Common Listing, In-depth Detail, Comparative Analysis, Process Steps, Examples, Ranking or Priority, Summary, and Default. Each type has specific diversity requirements and evaluation standards.

    \vspace{0.4em}
    \noindent\textbf{OUTPUT FORMAT} \\
    Strictly follow this JSON format: \texttt{\{ "analysis": \{ "think": "", "pass": true/false \} \}}

    \vspace{0.4em}
    \noindent\textbf{INPUT DATA} \\
    Chapter Outline: \texttt{\{chapter\_outline\}} Draft: \texttt{\{draft\}}

\end{StrategyBox}


\begin{StrategyBox}[frameblue]{bgblue}{Knowledge Enrichment Prompt}
\refstepcounter{promptidx}
\label{pro:9}

    \vspace{0.1em}
    \noindent\textbf{ROLE} \\
    You are a professional and detail-oriented information analyst, adept at synthesizing insights from multiple sources and clearly identifying their origins. Based on the following user query and knowledge excerpts, generate an accurate, well-structured, and source-traceable response that helps the user grasp the key conclusions. Automatically detect the user's primary language and ensure all responses are in that language.

    \vspace{0.4em}
    \noindent\textbf{INPUT DATA} \\
    \texttt{<chapter\_outline> \{chapter\_outline\} </chapter\_outline> <Known Perspectives and Knowledge> \{knowledge\} </Known Perspectives and Knowledge>}

    \vspace{0.4em}
    \noindent\textbf{GENERATION RULES} \\
    1. The response must remain closely aligned with the user query. Use clear and precise language, avoiding vagueness, redundancy, or circular phrasing. 2. You may integrate information from multiple excerpts but must not infer or speculate beyond what is explicitly provided. 3. Organize the response into several paragraphs if needed, each addressing a distinct fact or dimension. 4. Do not copy or list document contents verbatim. Instead, reorganize, summarize, and refine the language for clarity and cohesion. 5. Write in a natural, fluent style suitable for end users—avoid overly academic or mechanical phrasing. 6. Do not mention "document numbers" or "indexes." Source traceability should appear only through the \texttt{quote\_ids} field.

    \vspace{0.4em}
    \noindent\textbf{OUTPUT FORMAT} \\
    Please produce the final response according to the above requirements. Strictly follow this JSON structure: \texttt{\{ "answer": "", "quote\_ids": [""] \}}

\end{StrategyBox}


\begin{StrategyBox}[framegreen]{bggreen}{Content Generation System Prompt}
\refstepcounter{promptidx}
\label{pro:10}

    \vspace{0.1em}
    \noindent\textbf{ROLE} \\
    You are a report-writing expert in the \texttt{\{domain\}} field. Follow the rules and standards below strictly to produce content that is factually accurate, logically rigorous, coherent, and insightful. Automatically detect the user's primary language and ensure all responses are in that language.

    \vspace{0.4em}
    \noindent\textbf{CORE CONSTRAINTS} \\
    1. Truth First: Use only factual data from the "Reference Materials." Do not fabricate or introduce external information. 2. Precise Citation: Each argument (data, opinion, conclusion) must cite the reference number \texttt{[\textasciicircum num]} at the end of the sentence. When continuously citing the same source, mark only the last sentence. 3. Entity Matching: Data must correspond exactly to the correct entity. Cross-entity references are forbidden. 4. Focus on the Question: Stay strictly aligned with the user's core topic; avoid deviation.

    \vspace{0.4em}
    \noindent\textbf{WRITING STANDARDS} \\
    1. Logical Rigor: Each paragraph should focus on one central argument, supported by facts and data. Avoid fragmented listing. Evidence must be specific and directly support the argument. Do not generalize from a single case, and do not reuse the same fact in multiple arguments. Ensure the reasoning chain is complete and clear. Common structures include: Explanatory: phenomenon $\to$ cause $\to$ mechanism $\to$ impact $\to$ conclusion; Decision-making: need $\to$ options $\to$ evaluation $\to$ comparison $\to$ recommendation; Evaluation: standard $\to$ performance $\to$ comparison $\to$ judgment $\to$ conclusion; Predictive: foundation $\to$ trend $\to$ driver $\to$ scenario $\to$ forecast. Maintain natural transitions between paragraphs and sentences, using linking phrases like "further analysis shows," "this indicates," "by comparison," etc.

    \noindent 2. Depth and Insight: Analyze causal mechanisms rather than merely describing phenomena. Integrate multiple perspectives, including market, user, policy, and technology dimensions. Based on verified facts, make reasonable trend projections or outlooks without speculation.

    \noindent 3. Expression Standards: Highlight key data, conclusions, trends, and pain points in bold. Maintain objectivity and precision; use clear, concise language and avoid empty or colloquial expressions. Define technical terms or abbreviations at first mention; ensure writing style matches the report type (industry research / investment report / blog). Keep paragraph lengths relatively balanced.

    \vspace{0.4em}
    \noindent\textbf{USE OF VISUAL TOOLS} \\
    Use the following tools flexibly to improve clarity and readability. Chart Generation: Generate ECharts charts for visualizing data trends or relationships. Format: \texttt{<chart><description>}Explain the role of the chart in the text and specify the data dimensions\texttt{</description></chart>}. Table Generation: Used for presenting precise data and multi-dimensional comparisons (e.g., financial indicators, parameter comparisons, itemized lists). Format: \texttt{<table><title>}Table Title\texttt{</title><markdown>}Table content (in Markdown format)\texttt{</markdown></table>}. Execution Principles: 1. All charts must be generated strictly from the reference materials. Remove incomplete or invalid dimensions before supplementing missing data. 2. Follow the specified XML format for all tool calls; all unspecified parameters are considered mandatory.

\end{StrategyBox}


\begin{StrategyBox}[framegreen]{bggreen}{Content Generation User Prompt}
\refstepcounter{promptidx}
\label{pro:11}

    \vspace{0.1em}
    \noindent\textbf{TASK} \\
    Based on the "Reference", continue writing this chapter. Ensure logical consistency, formal expression, and natural connection with the previous text. Report creation time: \texttt{\{now\}} (prioritize the most recent and thematically relevant references).

    \vspace{0.4em}
    \noindent\textbf{WORKFLOW} \\
    Interpret Intent: Clearly identify the main subject, conditions, and focus of the user's question. Locate Evidence: Extract information from the "Reference Materials" closely related to the chapter outline. Write Content: Each paragraph should focus on a single argument with logical progression. Avoid reusing evidence. Non-summary sections should not end with summaries. Quality Check: Verify factual accuracy, citation consistency, logical soundness, and the sufficiency of evidence line by line.

    \vspace{0.4em}
    \noindent\textbf{CONTEXT INFORMATION} \\
    \texttt{<user\_query> \{query\} </user\_query> <chapter\_outline> \{chapter\_outline\} </chapter\_outline> <previous\_summary> \{above\} </previous\_summary> <outline> \{outline\} </outline> <reference> \{reference\} </reference>}

    \vspace{0.4em}
    \noindent\textbf{CONSTRAINTS} \\
    1. All data and facts must come directly from the reference materials. Fabrication or cross-entity use is prohibited. 2. Follow the chapter outline hierarchy. If no subheadings exist, output only the main body text without adding new levels. 3. Do not output any prompts, notes, or explanations.

\end{StrategyBox}

\begin{StrategyBox}[frameorange]{bgorange}{Intent Clarification Case}
\refstepcounter{caseidx}\label{case:1}
To provide a precise comparison, could you clarify:\\
1. Which LLM training scenario---large-scale foundational models, fine-tuning of midsize models, or research prototyping---are you focusing on?\\
2. Are you prioritizing raw throughput, total cost of ownership, or software ecosystem maturity?\\
3. Should the analysis include availability, supply-chain constraints, or projected 2025 price trends?
\end{StrategyBox}

\begin{StrategyBox}[frameorange]{bgorange}{Outline Case}
\refstepcounter{caseidx}\label{case:2}
\textbf{Title.} NVIDIA H100 vs. AMD MI300X: Comparative Analysis for Large-Scale Foundational Model Training in 2025.\\
\textbf{I. Industry Overview.} Evaluates the AI accelerator industry through PEST, focusing on export controls, cloud capex, generative-AI adoption, and transformer evolution; it identifies late-stage growth and the drivers of model complexity and energy efficiency.\\
\textbf{II. Market Landscape.} Quantifies market size and competitive dynamics for massive training workloads, including 2024--2025 projections, hyperscaler adoption, NVLink versus CDNA~3.0, cooling and interconnects, and deployment SWOT.\\
\textbf{III. Leading Players Analysis.} Benchmarks NVIDIA's CUDA, DGX, and CoWoS advantages against AMD's ROCm, open-standard, and TCO-focused approach, then compares throughput, energy efficiency, and memory architecture.\\
\textbf{IV. Industry Outlook.} Forecasts 2025 price-performance and procurement shifts and develops workload-specific decision matrices and TCO scenarios.\\
\textbf{V. Conclusion.} Strategic infrastructure choices balance NVIDIA's ecosystem maturity with AMD's memory-bandwidth advantages and supply-chain resilience.
\end{StrategyBox}

\end{document}